\documentclass[onecolumn]{fairmeta}
\usepackage[most]{tcolorbox}

\usepackage{amsmath}
\usepackage{amssymb}

\title{\algoname{}: Towards Functional Tool-Use Generalization via Keypoint Trajectory Reasoning}

\author[1]{Chuhao Zhou}
\author[2]{Liquan Wang}
\author[2]{Shuxin Cao}
\author[1]{Xiangyu Chen}
\author[1]{Yuxuan Hu}
\author[1]{Boyu Ma}
\author[2]{Animesh Garg}
\author[1, \dagger]{Jianfei Yang}

\affiliation[1]{MARS Lab, Nanyang Technological University}
\affiliation[2]{Georgia Institute of Technology}

\contribution[\dagger]{Corresponding Author}

\usepackage{graphicx}
\usepackage{amsmath}
\usepackage{amsfonts}
\usepackage{booktabs}
\usepackage{multirow} 
\usepackage[table]{xcolor}
\usepackage{wrapfig}
\usepackage{caption}
\usepackage{placeins}

\usepackage{xcolor}
\usepackage{enumitem}
\definecolor{cPink}{RGB}{255,105,180}

\newcommand{\algoname}{\textsc{FuncBridge}}

\abstract{
While humans readily repurpose a book, a stone, or a shoe to drive a nail, robots trained on specific tools fail to transfer the same function to novel ones -- a gap we formalize as \textit{functional generalization}.
Functionally equivalent tools share visually recognizable functional intent, such as where contact can occur and how a contact region should move to the target. However, this perceptual similarity does not directly carry over to action space, where each tool demands a different motor pattern to realize the function.
To bridge this gap, we explore intermediate representations including affordance images, human video prompts, functional videos and object masks, and 2D keypoint trajectories, finding that keypoint trajectories best balance functional expressiveness and action groundability. 
Building on this, {we present \algoname{}, a two-stage framework} that decouples functional reasoning from action execution: learning to predict generalizable keypoint trajectories from action-free data, then grounding them into robot actions with limited demonstrations. 
Across a benchmark spanning ten tools and three functions, including hitting, sweeping, and hooking, \algoname{} consistently outperforms state-of-the-art methods on unseen tools in both simulation and the real world.
}

\correspondence{Jianfei Yang at \email{jianfei.yang@ntu.edu.sg}}
\metadata[Project Page]{\url{https://chuhaozhou99.github.io/FuncBridge/}}

\begin{document}
\maketitle


\section{Introduction}
\label{sec:intro}
	 
In open-ended real-world environments, the right tool is rarely at hand, yet humans adapt effortlessly by repurposing available objects to accomplish functions. For example, in hitting, a book, a stone, or a shoe can be used to drive a nail because functionally equivalent tools share a common functional intent: a contact region to strike with and a motion that brings it onto the target. 
This capacity, which we term \textit{functional generalization}, is a hallmark of human dexterity that robots have yet to achieve. Current manipulation policies overfit to the appearance and geometry of seen tools, failing when handed a novel one that serves the same function~\citep{chen2025tool, turpin2021gift, qin2023robot, jia2026action, bai2026flash}.

\begin{figure}[ht]
    \centering
    \includegraphics[width=1.0\linewidth]{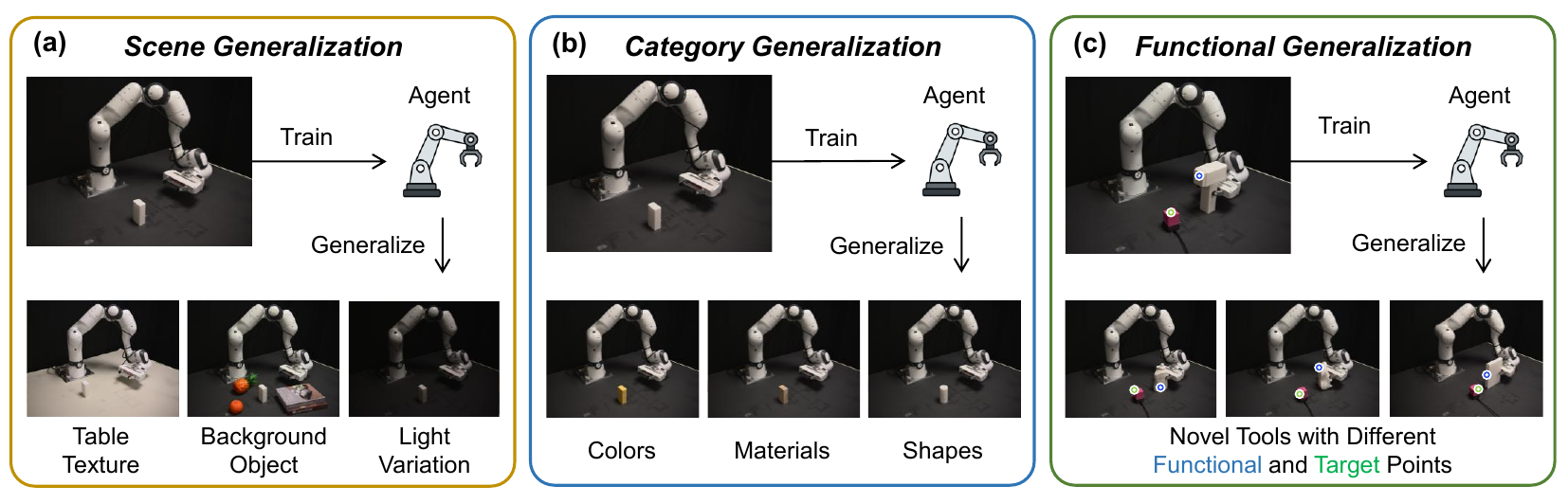}
    \caption{
    \textbf{Comparison of functional generalization with existing generalization settings.}
    (a) Scene generalization handles visual variations.
    (b) Category generalization transfers across object properties.
    (c) Functional generalization uses unseen tools for the same function.
    }
    \label{fig:functional_generalization}
\end{figure}
Functional generalization is fundamentally harder than conventional generalization. As illustrated in Fig.~\ref{fig:functional_generalization}, scene- and category-level generalization require tolerating visual variations while the underlying motion stays the same~\citep{goyal2023rvt, shridhar2023perceiver, shridhar2022cliport, nair2023r3m}. Functional generalization, by contrast, {demands adapting the motion to each tool's geometry}: striking a target with a novel tool requires locating its contact region, aligning it, and producing an appropriate motion, even when the tool's shape and required trajectory differ from training. {The core difficulty is a mismatch: functionally equivalent tools exhibit common functional intent that can be perceived from visual observations, but realizing the same function requires different motions across tools.}

Bridging this perception-to-action gap requires an intermediate representation that carries functional intent across tools, satisfying two competing demands: {it should be expressive enough to capture where contact can occur and how a selected contact region should move toward the target, yet abstract enough to discard tool-specific appearance that may hinder generalization to unseen tools.}
Dense representations like raw video entangle function-relevant cues with tool-specific appearance, while overly sparse ones discard the geometric and temporal structure needed for precise action grounding. This trade-off raises the central question of our work: \textit{which representation best transfers functional intent from perception to action?}

To answer this question, we evaluate intermediate representations 
including affordance images, human video prompts, {functional videos and object masks}, and 2D keypoint trajectories, 
finding that keypoint trajectories best balance functional expressiveness and 
action groundability: they capture function-relevant contact and motion structure 
while discarding tool-specific appearance that causes overfitting. Building on 
this, we propose \algoname{}, a two-stage framework that 
decouples functional reasoning from action execution.
{It learns a functional keypoint predictor from large-scale action-free data and an action-grounding policy from limited action-labeled demonstrations. At inference, the predictor predicts function-relevant keypoint trajectories, which the execution policy grounds into robot actions.
We further introduce a benchmark spanning ten tools and three functions, including hitting, sweeping, and hooking, on which \algoname{} consistently outperforms state-of-the-art methods on unseen tools in both simulation and the real world, achieving substantial gains in average success rate.}
In summary, our contributions are threefold:
\begin{itemize}[leftmargin=*]
    \item We formalize \textit{functional generalization} in robotic tool-use and identify the perception-to-action gap as its core difficulty.
    \item We find that 2D keypoint trajectories best balance functional 
    expressiveness and action groundability, and propose \algoname{}, {a two-stage framework that learns functional keypoint planning from action-free data and action grounding from limited robot demonstrations.}
   {\item Across ten tools and three functions in simulation, with real-world validation, we demonstrate that \algoname{} consistently outperforms state-of-the-art methods on unseen tools.}
\end{itemize}


\section{Related Works}
\label{sec:related_works}

\paragraph{Generalization Problems in Robotic Manipulation.}
Generalization has been a central problem in robotic manipulation, and existing studies mainly evaluate it from three perspectives: scene-level, category-level, and cross-embodiment generalization.
Scene-level generalization focuses on whether a policy can remain robust when the surrounding visual or physical contexts vary, such as changes in object layouts, lighting conditions, or camera viewpoints~\citep{goyal2023rvt, shridhar2023perceiver, shridhar2022cliport, nair2023r3m}. 
Category-level generalization studies whether robots can transfer manipulation skills to unseen objects that belong to the same semantic category while differing in geometry, size, or texture~\citep{gao2021kpam, mo2021where2act, geng2023gapartnet}.
More recently, cross-embodiment generalization has attracted attention, which aims to train a unified policy that can transfer across different robot platforms, such as robotic arms and humanoids~\citep{o2024open, team2024octo, intelligence2025pi}.
In this work, we go beyond generalization over visual appearance and robot embodiment and focus on \textit{functional generalization}, in which an agent is required to accomplish the same function using diverse, unseen tools.

\paragraph{Improving Generalization in Robot Policy Learning.}
To improve manipulation generalization, the most straightforward way is scaling up action-labeled robotic data and training policies end-to-end~\citep{zitkovich2023rt, o2024open, team2024octo, khazatsky2024droid, jia2026physics}. 
However, collecting large-scale robotic data is time-consuming and labor-intensive, especially for complicated tool-use manipulations. 
Reducing the dependence on action labels, some works learn transferable motion priors from large-scale action-free video data~\citep{nair2023r3m, wang2023mimicplay, wu2024unleashing,ma2026dit4dit, pai2025mimic, an2026feedback, hou2026world}. 
For example, MimicPlay~\citep{wang2023mimicplay} learns high-level latent plans from human play videos and grounds them into robot actions with a low-level visuomotor policy. 
Although videos contain rich motion and interaction cues, dense future frames do not explicitly reveal a compact abstraction of function-relevant intent. As a result, effectively learning functional priors from videos often requires extremely extensive pretraining.
Another line of work improves action grounding with more structured intermediate representations, such as interaction heatmaps~\citep{mo2021where2act, bahl2023affordances, wu2025afforddp, li2026compassad}, task-oriented grasps~\citep{fang2020tog}, object-centric representations~\citep{zhulearning, chen2025tool, chapin2025object, cao2026keygen}, or keypoint trajectories~\citep{gao2021kpam, huang2025rekep, turpin2021gift, wen2023any, cao2026keygen}.
For instance, ATM~\citep{wen2023any} predicts future trajectories of arbitrary points from video and uses them to guide a low-level policy for action generation. 
However, these methods mainly focus on standard manipulation tasks, such as object pick-up, while overlooking tool-use scenarios that require capturing more abstract functional intent, which are systematically investigated in our work.

\begin{figure}[ht]
    \centering
    \includegraphics[width=0.95\linewidth]{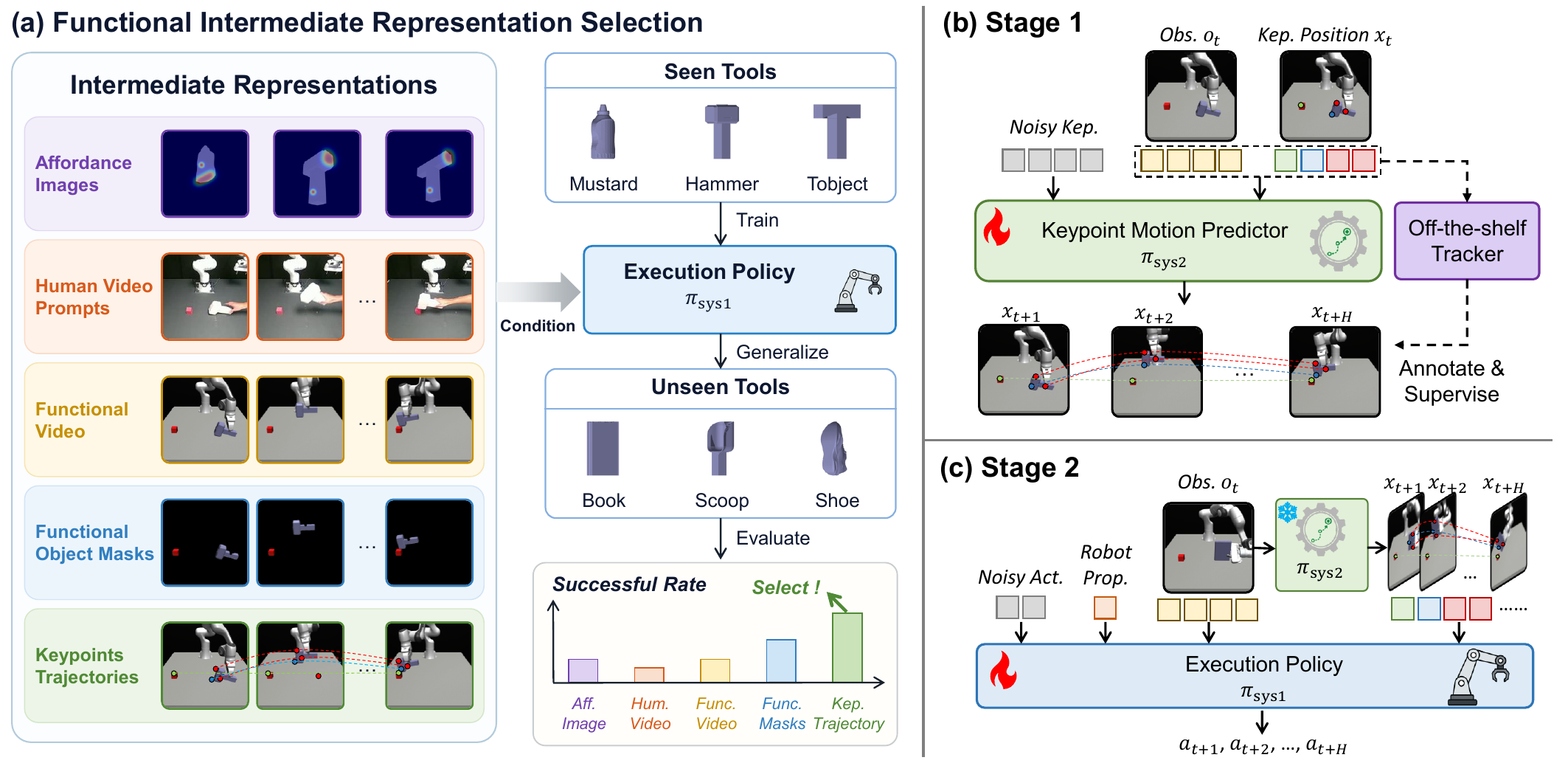}
    \caption{
    \textbf{Overview of the \algoname{}.} (a) Evaluation and selection of functional intermediate representations. (b) In Stage 1, the System-2 planner learns to predict future keypoint-based functional plans from action-free observations. (c) In Stage 2, the System-1 execution policy grounds the predicted functional plans into executable robot actions.
    }
    \label{fig:functional_generalization_pipeline}
\end{figure}

\vspace{-15pt}
\section{Methods}
\label{sec:method}

\subsection{Problem Formulation}
We formulate functional generalization in tool-use manipulation as a Markov Decision Process (MDP), where the state $\mathbf{s}_t$ consists of the visual observation $\mathbf{o}_t$, the proprioceptive state $\mathbf{q}_t$, and optionally a functional intermediate representation $\mathbf{x}_t$ (e.g., keypoint trajectories) that carries function-relevant cues bridging perception and action, and the action $\mathbf{a}_t$ is the motor command. A task function (e.g., hitting a target) is defined over seen tools $\mathcal{T}_{\text{seen}}$ for training and disjoint unseen tools $\mathcal{T}_{\text{unseen}}$ for evaluation, where the function stays fixed but the tool, its appearance, and the required contact positions and motions all change.

Two types of data are available for learning. The first is an action-free dataset $\mathcal{D}_{U}=\{(\mathbf{O}^{(i)}, \mathbf{X}^{(i)})\}_{i=1}^{N_U}$, consisting of visual observations $\mathbf{O}^{(i)}=\{\mathbf{o}^{(i)}_t\}_{t=1}^{T}$ and functional intermediate representations $\mathbf{X}^{(i)}=\{\mathbf{x}^{(i)}_t\}_{t=1}^{T}$, which is abundant and can span a broader range of tools since it requires no robot labels.
The second is a smaller action-labeled robotic dataset $\mathcal{D}_{L}=\{(\mathbf{O}^{(i)}, \mathbf{X}^{(i)}, \mathbf{Q}^{(i)}, \mathbf{A}^{(i)})\}_{i=1}^{N_L}$, which provides proprioceptive states $\mathbf{Q}^{(i)}$ and ground-truth actions $\mathbf{A}^{(i)}$ but is costly to collect, so $|\mathcal{D}_L| \ll |\mathcal{D}_U|$. This data asymmetry, together with the perception-to-action gap discussed in Sec.~\ref{sec:intro}, motivates a two-stage decomposition: 
\begin{equation}
\underbrace{
\pi_{\mathrm{FuncBridge}}(
\mathbf{a}_{t:t+H} \mid \mathbf{o}_{t}, \mathbf{q}_{t}, \mathbf{x}_{t}
)
}_{\text{FuncBridge}}=
\underbrace{
\pi_{\mathrm{sys2}}(
\hat{\mathbf{X}}_{t:t+H} \mid \mathbf{o}_{t}, \mathbf{x}_{t}
)
}_{\text{functional reasoning}}
\underbrace{
\pi_{\mathrm{sys1}}(
\mathbf{a}_{t:t+H} \mid \mathbf{o}_{t}, \mathbf{q}_{t}, \hat{\mathbf{X}}_{t:t+H}
)
}_{\text{grounded execution}} .
\label{eq:forge_decomposition}
\end{equation}
\algoname{} learns a future predictor on $\mathcal{D}_U$ for forecasting functional intermediate representations from observations, and an execution policy on $\mathcal{D}_L$ for grounding representations into actions.


\subsection{Functional Intermediate Representations Selection}
\label{sec:Functional Intermediate Representations Selection}
As discussed in Sec.~\ref{sec:intro}, a desired functional intermediate representation $\mathbf{X}_{t:t+H}$ should preserve the contact and motion structure shared across tools, while remaining predictable from action-free observations and groundable into precise robot actions. 
We hypothesize that 2D keypoint trajectories form a suitable choice of $\mathbf{X}_{t:t+H}$. 
They discard tool-specific appearance that may cause overfitting, while retaining compact geometric information such as the contact point, target position, and affordance-relevant tool structure. 
At the same time, they provide a structured temporal description of how function-relevant points should move, which makes them easy to predict from existing vision models~\citep{wen2023any} and to ground into actions. 
{To validate this hypothesis, we compare keypoint trajectories with four alternative choices of $\mathbf{X}_{t:t+H}$ spanning different temporal forms as shown in Fig.~\ref{fig:functional_generalization_pipeline}: affordance images as static conditions, human video prompts as pooled conditions, and functional videos and object masks as sequential conditions.}

\textbf{Affordance Image.}~~
Given an affordance image $I_{\mathrm{aff}}^{(i)}$, we encode it with an image encoder $E_{\mathrm{img}}$~\citep{he2016deep} and use the resulting feature as a time-invariant condition: $x_t^{(i)}=E_{\mathrm{img}}(I_{\mathrm{aff}}^{(i)})\in\mathbb{R}^{d}$.

\textbf{Human Video Prompt.}~~
Given a human video $V_{\mathrm{hum}}^{(i)}=\{I_{\mathrm{hum},t}^{(i)}\}_{t=1}^{T}$, we encode it with a video encoder $E_{\mathrm{vid}}$~\citep{bertasius2021space} and apply temporal average pooling to obtain a sequence-level feature, which is used as a time-invariant condition: $x_t^{(i)}=\mathrm{AvgPool}(E_{\mathrm{vid}}(V_{\mathrm{hum}}^{(i)}))\in\mathbb{R}^{d}$.

{\textbf{Functional Videos and Object Masks.}~~
Given a ground-truth functional execution video $V_{\mathrm{func}}^{(i)}=\{I_{\mathrm{func},t}^{(i)}\}_{t=1}^{T}$, we encode each frame with an image encoder $E_{\mathrm{img}}$, yielding a temporally varying condition $x_t^{(i)}=E_{\mathrm{img}}(I_{\mathrm{func},t}^{(i)})\in\mathbb{R}^{d}$. For object masks, as shown in Fig.~\ref{fig:functional_generalization_pipeline}, we remove the background while retaining the tool and target object in each frame, with the masked frames encoded similarly.}

\textbf{Keypoint Trajectory.}~~
We denote the keypoint trajectory as $P^{(i)}=\{p_t^{(i)}\}_{t=1}^{T}$, where each frame $p_t^{(i)}=\{p_t^{(i),k}\}_{k=1}^{K}$ contains $K$ 2D keypoints.
At time step $t$, we flatten the keypoint coordinates and encode them with a keypoint encoder $E_{\mathrm{kp}}$, yielding $x_t^{(i)}=E_{\mathrm{kp}}(\mathrm{Flatten}(p_t^{(i)}))\in\mathbb{R}^{d}$.

To evaluate these candidates, we conduct a provided-reference experiment in Fig.~\ref{fig:functional_generalization_pipeline}~(a), where each representation is pre-extracted for both seen and unseen tools and used as the condition for the execution policy $\pi_{\mathrm{sys1}}$. The policy is trained on action-labeled data from seen tools in $\mathcal{D}_{L}$ and evaluated on unseen tools for hitting, where a rollout is successful if the designated hitting region strikes the target. Among candidates, 2D keypoint trajectories achieve the best performance, since they go beyond static region cues and dense visual motion by providing a compact, structured description of how function-relevant points move and align over time. We therefore adopt 2D keypoint trajectories as the functional intermediate representation for \algoname{}.

\subsection{\algoname{}: Two-Stage Functional Reasoning and Grounded Execution}
\label{sec:FORGE}
After selecting 2D keypoint trajectories as the functional intermediate representation, we design a two-stage framework to decouple functional reasoning from action execution.
This enables \algoname{} to learn transferable functional keypoint motion from action-free demonstrations, while grounding the predicted trajectories into robot actions with limited action-labeled data.

\textbf{Stage 1: Action-free Functional Reasoning.}~~ 
As shown in Fig.~\ref{fig:functional_generalization_pipeline}~(b), we train the keypoint motion predictor $\pi_{\mathrm{sys2}}$ on the action-free dataset $\mathcal{D}_{U}$ to predict future functional keypoint trajectories. 
Given the visual observation $\mathbf{o}_t$ and keypoint coordinates $\mathbf{x}_t$ at time-step $t$, $\pi_{\mathrm{sys2}}$  generates a keypoint trajectory $\hat{\mathbf{X}}_{t:t+H}$ over horizon $H$. 

Formally, we instantiate $\pi_{\mathrm{sys2}}$ as a conditional flow-matching model in the keypoint space.
Given a ground-truth keypoint trajectory $\mathbf{X}_{t:t+H}$, a Gaussian noise sample $\epsilon\sim\mathcal{N}(0,\mathbf{I})$, and an interpolation time $\tau\sim\mathcal{U}(0,1)$, we define the interpolated keypoint trajectory as
$\mathbf{X}_{t:t+H}^{\tau}=\tau \mathbf{X}_{t:t+H} + (1-\tau)\epsilon
$.
The keypoints motion predictor learns a conditional velocity field $u_{\phi}$ by minimizing
\begin{equation}
    \mathcal{L}_{\mathrm{sys2}}
    =
    \mathbb{E}_{\tau,\epsilon,\mathcal{D}_{U}}
    \left[
    \left\|
    u_{\phi}
    \left(
    \mathbf{X}_{t:t+H}^{\tau}, \tau
    \mid
    \mathbf{o}_t, \mathbf{x}_t
    \right)
    -
    \left(
    \mathbf{X}_{t:t+H}-\epsilon
    \right)
    \right\|_2^2
    \right].
    \label{eq:sys2_flow_matching}
\end{equation}
During training, our task specifies the functional point on the tool and the target point on the object. 
Given the functional point, we use SAM2~\citep{ravi2025sam} to segment the tool mask and sample 5 keypoints on the mask with Farthest Point Sampling (FPS), resulting in 7 keypoints. 
As shown in Fig.~\ref{fig:functional_generalization_pipeline}~(b), we track these keypoints with CoTracker~\citep{karaev2024cotracker} to obtain keypoint trajectories for training. 
Since this stage only requires visual observations and keypoint trajectories, System-2 can leverage action-free data to capture transferable functional intent as keypoint-based motions, without relying on large-scale robot action labels.

\textbf{Stage 2: Grounded Execution Policy.}~~
In the second stage, we freeze the pretrained $\pi_{\mathrm{sys2}}$ and train the execution policy $\pi_{\mathrm{sys1}}$ on the action-labeled robotic dataset $\mathcal{D}_{L}$. 
Given the observation $\mathbf{o}_t$, robot state $\mathbf{q}_t$, and 2D keypoint coordinates $\mathbf{x}_t$, the frozen $\pi_{\mathrm{sys2}}$ first predicts future keypoint motions:
\begin{equation}
    \hat{\mathbf{X}}_{t:t+H}
    =
    \pi_{\mathrm{sys2}}(\mathbf{o}_t, \mathbf{x}_t).
\end{equation}
To improve robustness to prediction errors, we apply random pixel perturbations to the $\hat{\mathbf{X}}_{t:t+H}$ during training, where each keypoint is shifted by 1--5 pixels with a probability of 0.5. The execution policy $\pi_{\mathrm{sys1}}$ then grounds the perturbed keypoint trajectories $\tilde{\mathbf{X}}_{t:t+H}$ into executable robot actions:
\begin{equation}
    \hat{\mathbf{A}}_{t:t+H}
    =
    \pi_{\mathrm{sys1}}(\mathbf{o}_t, \mathbf{q}_t, \tilde{\mathbf{X}}_{t:t+H}).
\end{equation}

Similar to $\pi_{\mathrm{sys2}}$, we instantiate $\pi_{\mathrm{sys1}}$ as a conditional flow-matching model in the action space.
Given a ground-truth action chunk $\mathbf{A}_{t:t+H}$, a Gaussian noise sample $\epsilon\sim\mathcal{N}(0,\mathbf{I})$, and an interpolation time $\tau\sim\mathcal{U}(0,1)$, we define the interpolated action trajectory as
$
    \mathbf{A}_{t:t+H}^{\tau}
    =
    \tau \mathbf{A}_{t:t+H} + (1-\tau)\epsilon
$.
The action-grounded execution policy learns a conditional velocity field $v_{\theta}$ by minimizing:
\begin{equation}
    \mathcal{L}_{\mathrm{sys1}}
    =
    \mathbb{E}_{\tau,\epsilon,\mathcal{D}_{L}}
    \left[
    \left\|
    v_{\theta}
    \left(
    \mathbf{A}_{t:t+H}^{\tau}, \tau
    \mid
    \mathbf{o}_t, \mathbf{q}_t, \mathbf{\tilde{X}}_{t:t+H}
    \right)
    -
    \left(
    \mathbf{A}_{t:t+H}-\epsilon
    \right)
    \right\|_2^2
    \right].
    \label{eq:sys1_flow_matching}
\end{equation}


\begin{figure}[t]
    \centering
    \includegraphics[width=0.95\linewidth]{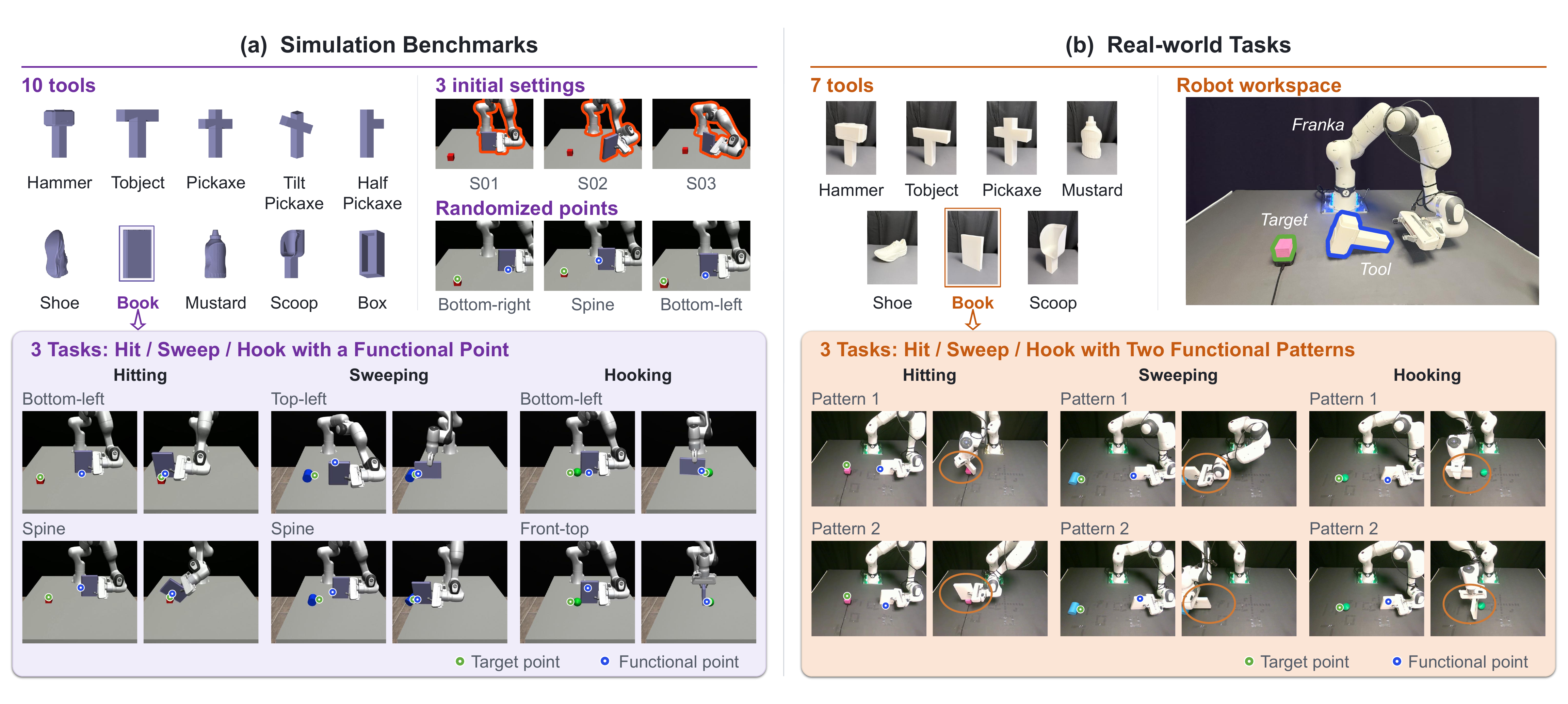}
    \caption{
   \textbf{Simulation Benchmark and Real-World Setting.} (a) The simulation benchmark contains 10 tools and three functions, including hitting, sweeping, and hooking, with multiple initial settings and randomized functional and target points. (b) The real-world benchmark uses a Franka robot with 7 tools and the same three functions, each evaluated under two functional patterns.}
    \label{fig:experiment_settings}
\end{figure}

\vspace{-15pt}
\section{{Experiments}}
\label{sec:experiments}


Functional generalization remains largely underexplored in robotic manipulation.
In this work, we instantiate it across three representative tool-use functions, including hitting, sweeping, and hooking, and construct both simulation and real-world benchmarks to evaluate whether a policy can transfer each function to unseen tools with diverse geometries.
Specifically, we test four hypotheses underlying the central claims of this work:

\begin{itemize}[leftmargin=*]
    \item \textbf{H1}: Functional generalization benefits from an explicit, function-aware intermediate representation rather than direct perception-to-action mapping or generic motion cues.
    \item \textbf{H2}: Among candidate intermediate representations, 2D keypoint trajectories best balance functional expressiveness and action groundability.
    
    \item \textbf{H3}: Functional generalization can be achieved with moderate action-labeled tool diversity.
    
    \item \textbf{H4}: Functional reasoning is scalable, which can benefit from more diverse action-free data and demonstrate greater autonomy through VLM-based functional point selection.
\end{itemize}

\subsection{Experimental setup}
\label{sec:experiments_setup}

\textbf{Simulation Benchmark.}~~
As shown in Fig.~\ref{fig:experiment_settings}, our simulation benchmark covers 10 tools and three functions, including hitting, sweeping, and hooking, with three initial settings per tool.
For each tool-setting-function combination, we collect 30 demonstrations using function-specific MPPI planners, totaling 2,700 demonstrations.
In the main setting, all methods are trained on action-labeled data from four tools (hammer, tobject, pickaxe, and mustard) and evaluated on three unseen tools held out from action-labeled training (scoop, shoe, and book) across all functions.
Each unseen tool is evaluated under three initial settings with 30 rollouts per setting.
For each demonstration, we randomize the target position and designated functional point, with different points possible for the same tool and function (e.g., using the hammer head or handle for hitting).
A rollout is successful only if the specified function is completed using that point: \textit{hitting} requires the hitting point to strike the red cube, \textit{sweeping} requires the sweeping point to push the blue cylinder, and \textit{hooking} requires the hooking point to pull the green ball inward.



\textbf{Real-World Setting.}~~
We further construct a real-world benchmark on a Franka robot platform.
The benchmark covers 7 tools and the same three functions, with two functional patterns defined for each function.
For each training tool and functional pattern, we collect 20 expert demonstrations.
For hitting, the policies are trained on hammer, pickaxe, and mustard and evaluated on three unseen tools (tobject, book, and shoe).
For sweeping and hooking, they are trained on hammer and mustard and evaluated on the unseen book.
Each functional pattern is evaluated with 10 real-world rollouts.
Unlike simulation, the designated functional point specifies the required functional pattern rather than serving as a point-level success criterion. A rollout is successful if the robot completes the specified function following the corresponding pattern.


\textbf{Baselines.}~~
We compare \algoname{} against three baselines covering end-to-end and keypoint-based policies.
All baselines receive the same inputs as \algoname{}: visual observations, proprioceptive states, and the initial functional, target, and tool point positions.
\textit{Flow-Matching (FM)}~\citep{lipmanflow} and \textit{Diffusion Policy (DP)}~\citep{chi2025diffusion} are end-to-end visuomotor policies that map observations directly to actions, serving as references for the perception-to-action gap (Sec.~\ref{sec:intro}).
\textit{ATM}~\citep{wen2023any} conditions the execution policy on keypoint trajectories. Given the functional and target points, it uses its generic any-point trajectory model pretrained on LIBERO-90 to predict their trajectories. This serves as a keypoint-based baseline using generic trajectory modeling rather than explicitly predicting function-aware keypoint motion.

\textbf{Evaluation Metrics.}~~
All experiments report the success rate (SR) as the main evaluation metric, which is defined as the number of successful trials divided by the total number of rollout trials.

\begin{table}[ht]
\centering
\caption{\textbf{Simulation Results on Unseen Tools.} Success rate (SR) on three unseen tools across hitting, sweeping, and hooking. Each tool is evaluated under 3 settings with 30 rollouts per setting.}
\label{tab:simulation_results}
\small
\setlength{\tabcolsep}{3.0pt}

\resizebox{0.9\linewidth}{!}{
\begin{tabular}{lcccccccccccc}
\toprule
& \multicolumn{4}{c}{Hitting} 
& \multicolumn{4}{c}{Sweeping}
& \multicolumn{4}{c}{Hooking} \\
\cmidrule(lr){2-5}
\cmidrule(lr){6-9}
\cmidrule(lr){10-13}

Method 
& Scoop & Shoe & Book & Avg. 
& Scoop & Shoe & Book & Avg.
& Scoop & Shoe & Book & Avg. \\
\midrule

FM 
& 0.07 & 0.03 & 0.16 & \cellcolor{gray!15} 0.09 
& 0.49 & 0.23 & 0.19 & \cellcolor{gray!15} 0.30
& 0.22 & 0.06 & 0.14 & \cellcolor{gray!15} 0.14 \\

DP 
& 0.18 & 0.08 & 0.26 & \cellcolor{gray!15} 0.17 
& 0.49 & 0.18 & 0.13 & \cellcolor{gray!15} 0.27
& 0.12 & 0.06 & 0.08 & \cellcolor{gray!15} 0.09 \\

ATM 
& 0.12 & 0.06 & 0.06 & \cellcolor{gray!15} 0.08 
& 0.41 & 0.16 & 0.19 & \cellcolor{gray!15} 0.25
& 0.10 & 0.07 & 0.24 & \cellcolor{gray!15} 0.14 \\

\midrule

\algoname{} 
& \textbf{0.42} & \textbf{0.23} & \textbf{0.42} 
& \cellcolor{gray!15}\textbf{0.36} 
& \textbf{0.71} & \textbf{0.29} & \textbf{0.34} 
& \cellcolor{gray!15}\textbf{0.45}
& \textbf{0.29} & \textbf{0.17} & \textbf{0.24} 
& \cellcolor{gray!15}\textbf{0.23} \\

\bottomrule
\end{tabular}
}
\end{table}

\begin{table}[ht]
\centering
\caption{\textbf{Real-World Results on Unseen Tools.} Success rate (SR) under two functional patterns for hitting, sweeping, and hooking. Hitting is evaluated on three unseen tools, while sweeping and hooking are evaluated on the unseen book, with 10 rollouts per pattern.}

\label{tab:real_world_results}
\small
\setlength{\tabcolsep}{2.0pt}

\resizebox{0.9\linewidth}{!}{
\begin{tabular}{lccccccccccccccc}
\toprule
& \multicolumn{9}{c}{\textbf{Hitting}}
& \multicolumn{3}{c}{\textbf{Sweeping}}
& \multicolumn{3}{c}{\textbf{Hooking}} \\
\cmidrule(lr){2-10}
\cmidrule(lr){11-13}
\cmidrule(lr){14-16}

& \multicolumn{3}{c}{Tobject}
& \multicolumn{3}{c}{Book}
& \multicolumn{3}{c}{Shoe}
& \multicolumn{3}{c}{Book}
& \multicolumn{3}{c}{Book} \\
\cmidrule(lr){2-4}
\cmidrule(lr){5-7}
\cmidrule(lr){8-10}
\cmidrule(lr){11-13}
\cmidrule(lr){14-16}

Method
& P01 & P02 & Avg.
& P01 & P02 & Avg.
& P01 & P02 & Avg.
& P01 & P02 & Avg.
& P01 & P02 & Avg. \\
\midrule

FM
& 0.00 & 0.60 & \cellcolor{gray!15} 0.30
& 0.50 & 0.00 & \cellcolor{gray!15} 0.25
& 0.00 & 0.20 & \cellcolor{gray!15} 0.10
& 0.30 & 0.00 & \cellcolor{gray!15} 0.15
& 0.20 & 0.00 & \cellcolor{gray!15} 0.10 \\

\algoname{}
& \textbf{0.60} & \textbf{0.80} & \cellcolor{gray!15}\textbf{0.70}
& \textbf{0.60} & \textbf{1.00} & \cellcolor{gray!15}\textbf{0.80}
& \textbf{0.20} & \textbf{0.60} & \cellcolor{gray!15}\textbf{0.40}
& \textbf{0.80} & \textbf{0.30} & \cellcolor{gray!15}\textbf{0.55}
& \textbf{0.90} & \textbf{0.30} & \cellcolor{gray!15}\textbf{0.60} \\

\bottomrule
\end{tabular}
}

\end{table}

\subsection{Main Results \& Analysis}

\textbf{Simulation Results.}~~
As shown in Tab.~\ref{tab:simulation_results}, \algoname{} outperforms all baselines across hitting, sweeping, and hooking on unseen tools.
Compared with the best baseline results of $0.17$, $0.30$, and $0.14$, it achieves average SRs of $0.36$, $0.45$, and $0.23$, respectively.
The weak performance of FM and DP suggests that direct perception-to-action mapping struggles to adapt tool-specific motions to unseen geometries without tool-specific action supervision.
Despite being conditioned on the designated functional and target points, ATM performs less effectively, suggesting that its generic keypoint predictor may fail to produce effective function-relevant motion guidance for diverse tool geometries.
Together, these results support \textbf{H1}: explicit function-aware intermediate representation provides useful guidance for functional generalization.

\textbf{Real-World Results.}~~
We test \algoname{} on the real-world benchmark across all three functions.
As shown in Tab.~\ref{tab:real_world_results}, \algoname{} consistently outperforms the FM baseline across all evaluated unseen-tool and functional-pattern settings. These results extend the evidence for \textbf{H1} to real-world manipulation, showing that function-aware keypoint plans provide more reliable guidance.

\begin{figure}[ht]
    \centering
    \includegraphics[width=0.82\linewidth]{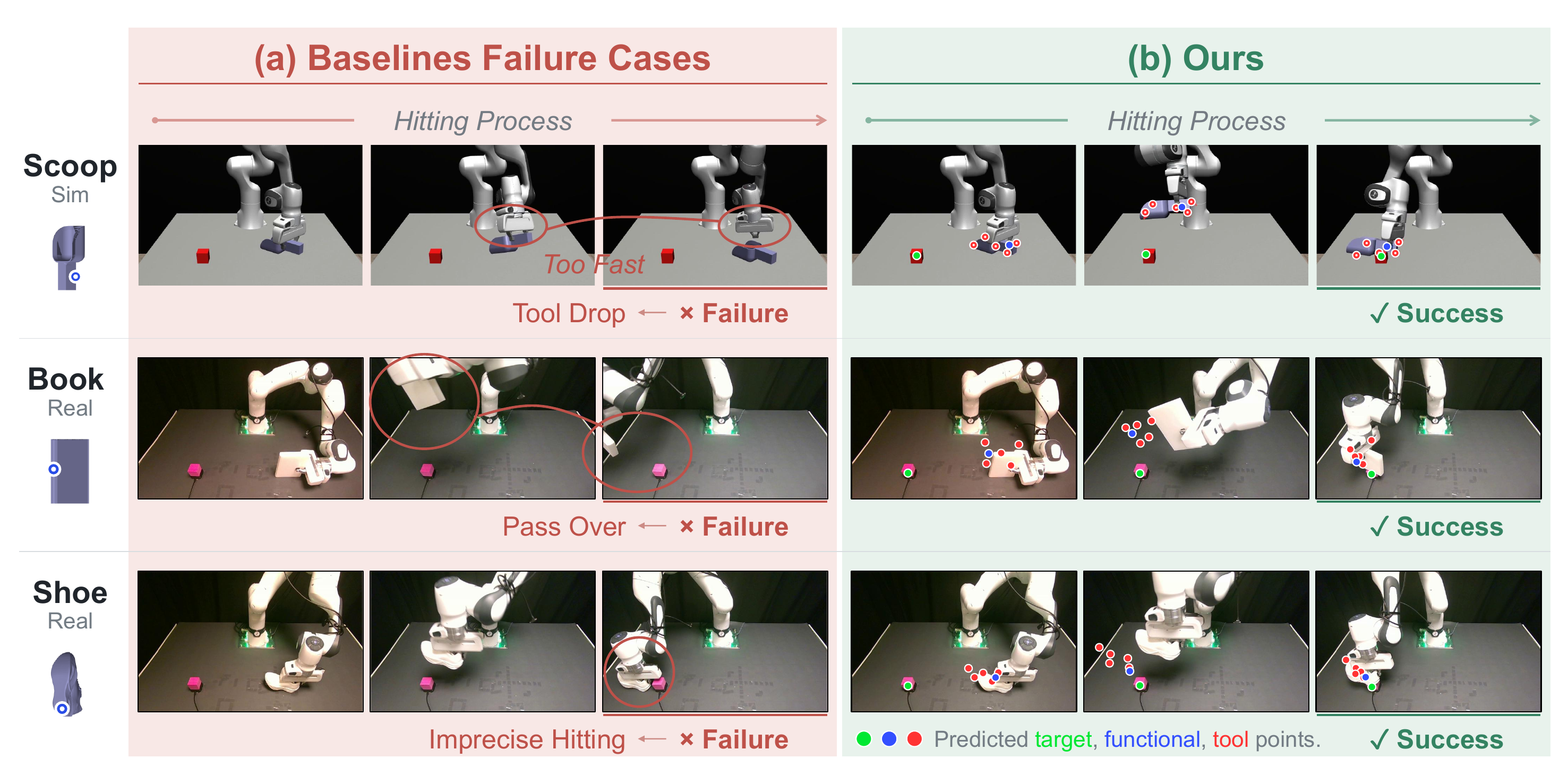}
    \caption{\textbf{Qualitative Comparison on Unseen Tools.} Representative baseline failure cases and successful rollouts of \algoname{} for hitting in simulation and the real world.}

    \label{fig:qualitative-analysis}
\end{figure}

\textbf{Failure Case Analysis.}~~
We further examine representative baseline failures on the hitting function in both simulation and the real world.
As shown in Fig.~\ref{fig:qualitative-analysis}, the baselines exhibit three typical failure modes: \emph{pass over}, where the tool moves across the target without establishing the intended contact; \emph{imprecise hitting}, caused by insufficient fine-grained spatial alignment; and \emph{tool dropping}, associated with unstable or overly aggressive motions.
In the corresponding cases, \algoname{} successfully completes the task by predicting function-aware keypoint trajectories that guide the functional region toward the target with the appropriate motion and alignment.
Additional qualitative comparisons across hitting, sweeping, and hooking in both simulation and the real world are provided in App.~\ref{app:additional_experimental_results}.

\subsection{Ablation Studies}
Beyond the main comparisons, we conduct two ablations on the key design choices of \algoname{}: the functional intermediate representation and the diversity of action-labeled tools.
Unless otherwise specified, all ablations are conducted on the hitting function in simulation.

\subsubsection{Effect of Functional Intermediate Representations.}

\begin{table}[ht]
\centering
\caption{\textbf{Ablation Results on Functional Intermediate Representations.} Success rate (SR) for hitting on unseen tools when different representations condition the same execution policy $\pi_{\mathrm{sys1}}$.}
\label{tab:ablation_intermediate_representation}
\small

\begin{tabular*}{0.9\linewidth}{
@{\extracolsep{\fill}}
llccc>{\columncolor{gray!12}}c
}
\toprule
Representation & Temporal input & Scoop & Shoe & Book & \cellcolor{white}Avg. \\
\midrule
None (FM) & -- & 0.07 & 0.03 & 0.16 & 0.09 \\
Affordance image & Static & 0.34 & 0.11 & 0.16 & 0.20 \\
Human video prompt & Pooled & 0.26 & 0.09 & 0.23 & 0.19 \\
\midrule
Per-frame video & Sequence & 0.28 & 0.13 & 0.29 & 0.23 \\
Per-frame object masks & Sequence & 0.37 & 0.16 & 0.43 & 0.32 \\
Keypoint trajectories & Sequence &
\textbf{0.50} & \textbf{0.31} & \textbf{0.67} & \textbf{0.49} \\
\bottomrule
\end{tabular*}

\end{table}

As discussed in Sec.~\ref{sec:Functional Intermediate Representations Selection}, we investigate the effect of functional intermediate representations by training and evaluating the execution policy $\pi_{\mathrm{sys1}}$ with pre-extracted representations as conditions.
Results in Tab.~\ref{tab:ablation_intermediate_representation} show that static affordance images and pooled human video prompts improve generalization, indicating that functional-region cues and motion demonstrations provide guidance.
Providing temporally varying representations improves performance, while object masks achieve stronger generalization by only retaining object-centric motion cues.
Among representations, 2D keypoint trajectories perform best, as they compactly encode fine-grained functional and target locations together with their structured motion over time.
These results support \textbf{H2}: keypoint trajectories best balance functional expressiveness and action groundability.

\subsubsection{Effect of Action-Labeled Tool Diversity.}
\begin{wraptable}{r}{0.54\linewidth}
\vspace{-0.8em}
\centering
\captionsetup{skip=4pt}
\caption{\textbf{Ablation on Action-Labeled Tool Diversity.} Success rate (SR) of \algoname{} when the execution policy $\pi_{\mathrm{sys1}}$ is trained with action-labeled data from $M=1 \text{,} 4 \text{,} 7$ tools and consistently evaluated on the same three unseen tools: scoop, shoe, and book.}
\label{tab:ablation_tool_generalization_setting}

\resizebox{0.95\linewidth}{!}{
\begin{tabular}{lcccc}
\toprule
\multirow{2}{*}{Method}
& \multicolumn{4}{c}{Tools} \\
\cmidrule(lr){2-5}
& Scoop & Shoe & Book & Avg. \\
\midrule
\algoname{} (1-to-3) & 0.14 & 0.10 & 0.23 & \cellcolor{gray!12} 0.16 \\
\algoname{} (4-to-3) & 0.42 & 0.23 & 0.42 & \cellcolor{gray!12} 0.36 \\
\algoname{} (7-to-3) & 0.39 & 0.27 & 0.52 & \cellcolor{gray!12} 0.39 \\
\bottomrule
\end{tabular}
}
\vspace{-0.8em}
\end{wraptable}
Beyond the standard 4-to-3 setting, we vary the number of action-labeled training tools while consistently evaluating on the same three unseen tools: scoop, shoe, and book.
As shown in Tab.~\ref{tab:ablation_tool_generalization_setting}, increasing the training diversity from 1 to 4 tools substantially improves the average SR from $0.16$ to $0.36$, while further increasing it to 7 tools yields a smaller gain to $0.39$.
These results support \textbf{H3}: functional generalization can be achieved with moderate action-labeled tool diversity in our two-stage framework.

\begin{figure}[ht]
    \centering
    \includegraphics[width=0.95\linewidth]{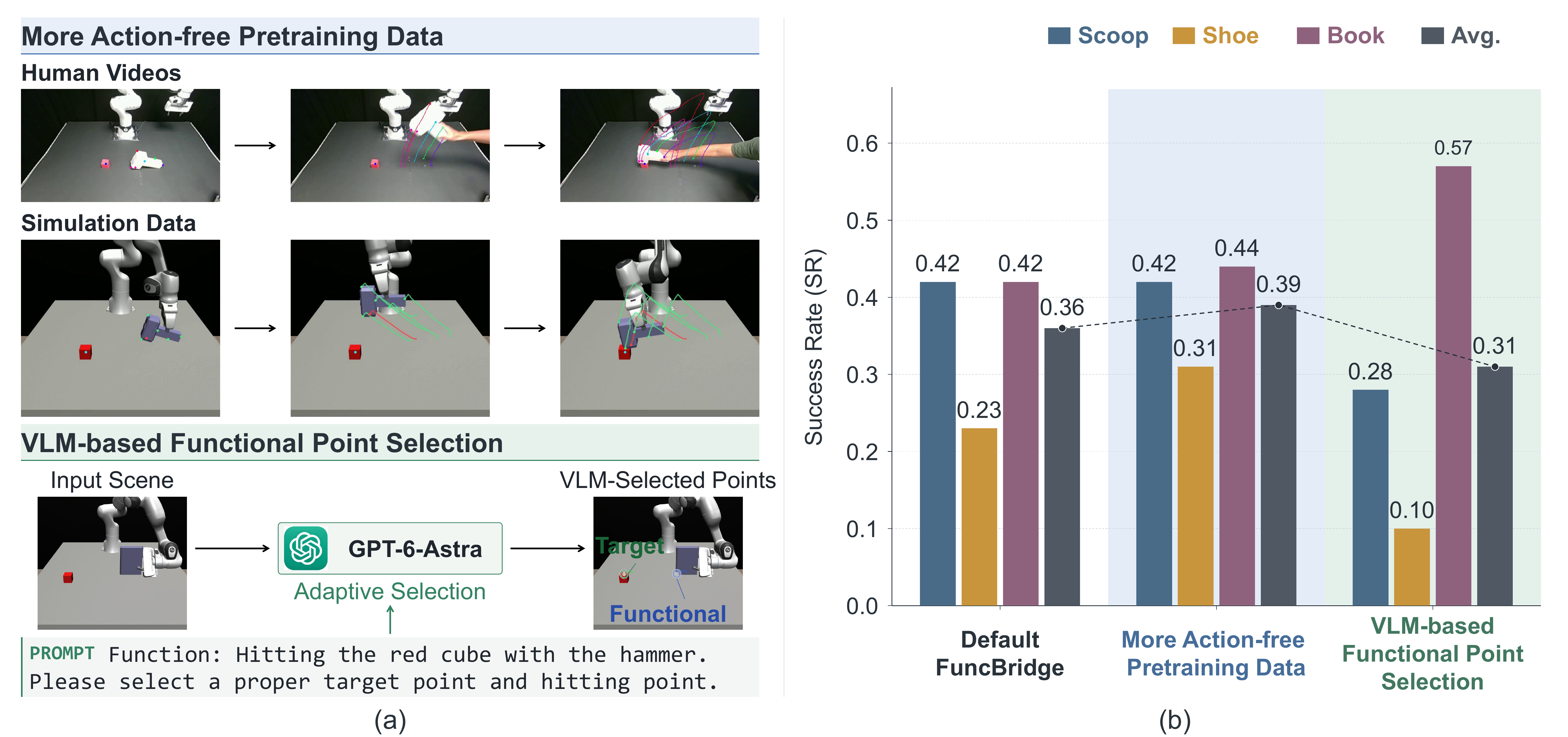}
    \caption{\textbf{Scalability of \algoname{}.} (a) Extensions using human videos and VLM-based functional point selection. (b) Success rates on unseen tools for the default model and each extension.}

    \label{fig:further-analysis}
\end{figure}





\subsection{Scalability of \algoname{}}
\label{sec:scalability}

Our default setting learns functional reasoning from action-free simulation videos, while the functional and target points are provided during evaluation.
We further examine \textbf{H4} by extending \algoname{} with multi-source action-free data and VLM-based function-relevant point selection.

\subsubsection{Scaling Action-Free Pretraining with Human Videos.}
\label{sec:scalability-human-video}
As shown in Fig.~\ref{fig:further-analysis} (a), we additionally collect and annotate keypoints of 525 human videos for the hitting function (see App.~\ref{app:fir_extraction} for details).
Together with the original action-free data, these videos are used to pretrain $\pi_{\mathrm{sys2}}$, while $\pi_{\mathrm{sys1}}$ is fine-tuned with the same four-tool action-labeled robotic data.
As shown in Fig.~\ref{fig:further-analysis} (b), incorporating human videos improves the average SR from $0.36$ to $0.39$.
This result shows that \algoname{} can benefit from heterogeneous action-free data, suggesting the potential of broader data sources for functional reasoning.

\subsubsection{Flexible VLM-Based Functional Point Selection.}
Given the observed scene and task instruction, GPT-6-Astra is prompted to adaptively select the functional point on the tool and the target point, which are then used by \algoname{} for functional planning and grounded execution.
A rollout is considered successful if the VLM-selected functional point is used to strike the target object, following the same hitting objective as the default setting.
As shown in Fig.~\ref{fig:further-analysis}, VLM-selected points achieve an average SR of $0.31$, compared with $0.36$ using manually specified points.
The performance gap mainly comes from VLM errors that select incorrect or unsuitable hitting points, such as the top side of a book.
Nevertheless, the results show that VLM-based point selection can be integrated with \algoname{} to reduce manual specification and support more autonomous functional tool use.

\section{Conclusion and Limitation}
\label{sec:conclusion}
We identify \textit{functional generalization} as an important yet underexplored problem, where an agent must accomplish the same function with unseen tools.
We introduce \algoname{}, a two-stage framework that decouples action-free functional reasoning from grounded execution. Across hitting, sweeping, and hooking, simulation and real-world experiments show improved generalization to unseen tools over existing baselines. Further analyses show that \algoname{} benefits from heterogeneous action-free data, including human videos, and can integrate VLM-based functional point selection for more autonomous tool use.

\textbf{Limitations and Future Work.}
\algoname{} remains limited in generalizing to unseen tools that require substantially different functional motions. Future work could leverage larger-scale egocentric or web videos with more diverse tool-use behaviors to further improve functional generalization. In addition, our method focuses on functional motion reasoning and does not explicitly model dexterous interaction, such as grasp adaptation or in-hand manipulation, which is also essential for general tool use and could be incorporated in future work.

\clearpage
\newpage
\bibliographystyle{assets/plainnat}
\bibliography{paper}

@article{qin2023robot,
  title={Robot tool use: A survey},
  author={Qin, Meiying and Brawer, Jake and Scassellati, Brian},
  journal={Frontiers in Robotics and AI},
  year={2023},
}

@inproceedings{goyal2023rvt,
  title={Rvt: Robotic view transformer for 3d object manipulation},
  author={Goyal, Ankit and Xu, Jie and Guo, Yijie and Blukis, Valts and Chao, Yu-Wei and Fox, Dieter},
  booktitle={Conference on Robot Learning},
  year={2023},
}

@inproceedings{shridhar2023perceiver,
  title={Perceiver-actor: A multi-task transformer for robotic manipulation},
  author={Shridhar, Mohit and Manuelli, Lucas and Fox, Dieter},
  booktitle={Conference on Robot Learning},
  year={2023},
}

@inproceedings{shridhar2022cliport,
  title={Cliport: What and where pathways for robotic manipulation},
  author={Shridhar, Mohit and Manuelli, Lucas and Fox, Dieter},
  booktitle={Conference on robot learning},
  year={2022},
}

@inproceedings{nair2023r3m,
  title={R3M: A Universal Visual Representation for Robot Manipulation},
  author={Nair, Suraj and Rajeswaran, Aravind and Kumar, Vikash and Finn, Chelsea and Gupta, Abhinav},
  booktitle={Conference on Robot Learning},
  year={2023},
}

@article{gao2021kpam,
  title={kpam 2.0: Feedback control for category-level robotic manipulation},
  author={Gao, Wei and Tedrake, Russ},
  journal={IEEE Robotics and Automation Letters},
  year={2021},
}

@inproceedings{mo2021where2act,
  title={Where2act: From pixels to actions for articulated 3d objects},
  author={Mo, Kaichun and Guibas, Leonidas J and Mukadam, Mustafa and Gupta, Abhinav and Tulsiani, Shubham},
  booktitle={Proceedings of the IEEE/CVF International Conference on Computer Vision},
  year={2021}
}

@inproceedings{geng2023gapartnet,
  title={Gapartnet: Cross-category domain-generalizable object perception and manipulation via generalizable and actionable parts},
  author={Geng, Haoran and Xu, Helin and Zhao, Chengyang and Xu, Chao and Yi, Li and Huang, Siyuan and Wang, He},
  booktitle={Proceedings of the IEEE/CVF conference on computer vision and pattern recognition},
  year={2023}
}

@article{fang2020tog,
  title   = {Learning Task-Oriented Grasping for Tool Manipulation from Simulated Self-Supervision},
  author  = {Fang, Kuan and Zhu, Yuke and Garg, Animesh and Kurenkov, Andrey and Mehta, Viraj and Fei-Fei, Li and Savarese, Silvio},
  journal = {The International Journal of Robotics Research},
  volume  = {39},
  number  = {2-3},
  pages   = {202--216},
  year    = {2020},
  doi     = {10.1177/0278364919872545}
}

@inproceedings{cao2026keygen,
  title     = {Key{G}en: Unsupervised Keypoint-Based Object-Centric Representations for Category-Level Policy Generalization},
  author    = {Cao, Shuxin and Wang, Liquan and Moghani, Masoud and Joffe, Benjamin and Garg, Animesh},
  booktitle = {IEEE/RSJ International Conference on Intelligent Robots and Systems (IROS)},
  year      = {2026}
}

@inproceedings{o2024open,
  title={Open x-embodiment: Robotic learning datasets and rt-x models: Open x-embodiment collaboration 0},
  author={O’Neill, Abby and Rehman, Abdul and Maddukuri, Abhiram and Gupta, Abhishek and Padalkar, Abhishek and Lee, Abraham and Pooley, Acorn and Gupta, Agrim and Mandlekar, Ajay and Jain, Ajinkya and others},
  booktitle={2024 IEEE International Conference on Robotics and Automation (ICRA)},
  year={2024},
}

@article{team2024octo,
  title={Octo: An open-source generalist robot policy},
  author={Team, Octo Model and Ghosh, Dibya and Walke, Homer and Pertsch, Karl and Black, Kevin and Mees, Oier and Dasari, Sudeep and Hejna, Joey and Kreiman, Tobias and Xu, Charles and others},
  journal={arXiv preprint arXiv:2405.12213},
  year={2024}
}

@article{intelligence2025pi,
  title={pi0.5: a Vision-Language-Action Model with Open-World Generalization},
  author={Intelligence, Physical and Black, Kevin and Brown, Noah and Darpinian, James and Dhabalia, Karan and Driess, Danny and Esmail, Adnan and Equi, Michael and Finn, Chelsea and Fusai, Niccolo and others},
  journal={arXiv preprint arXiv:2504.16054},
  year={2025}
}

@inproceedings{zitkovich2023rt,
  title={Rt-2: Vision-language-action models transfer web knowledge to robotic control},
  author={Zitkovich, Brianna and Yu, Tianhe and Xu, Sichun and Xu, Peng and Xiao, Ted and Xia, Fei and Wu, Jialin and Wohlhart, Paul and Welker, Stefan and Wahid, Ayzaan and others},
  booktitle={Conference on Robot Learning},
  pages={2165--2183},
  year={2023},
  organization={PMLR}
}

@article{khazatsky2024droid,
  title={Droid: A large-scale in-the-wild robot manipulation dataset},
  author={Khazatsky, Alexander and Pertsch, Karl and Nair, Suraj and Balakrishna, Ashwin and Dasari, Sudeep and Karamcheti, Siddharth and Nasiriany, Soroush and Srirama, Mohan Kumar and Chen, Lawrence Yunliang and Ellis, Kirsty and others},
  journal={arXiv preprint arXiv:2403.12945},
  year={2024}
}

@inproceedings{wang2023mimicplay,
  title={MimicPlay: Long-Horizon Imitation Learning by Watching Human Play},
  author={Wang, Chen and Fan, Linxi and Sun, Jiankai and Zhang, Ruohan and Fei-Fei, Li and Xu, Danfei and Zhu, Yuke and Anandkumar, Anima},
  booktitle={Conference on Robot Learning},
  year={2023},
}

@inproceedings{wu2024unleashing,
  title={Unleashing large-scale video generative pre-training for visual robot manipulation},
  author={Wu, Hongtao and Jing, Ya and Cheang, Chilam and Chen, Guangzeng and Xu, Jiafeng and Li, Xinghang and Liu, Minghuan and Li, Hang and Kong, Tao},
  booktitle={International Conference on Learning Representations},
  year={2024}
}

@article{ma2026dit4dit,
  title={Dit4dit: Jointly modeling video dynamics and actions for generalizable robot control},
  author={Ma, Teli and Zheng, Jia and Wang, Zifan and Jiang, Chunli and Cui, Andy and Liang, Junwei and Yang, Shuo},
  journal={arXiv preprint arXiv:2603.10448},
  year={2026}
}

@article{pai2025mimic,
  title={mimic-video: Video-action models for generalizable robot control beyond vlas},
  author={Pai, Jonas and Achenbach, Liam and Montesinos, Victoriano and Forrai, Benedek and Mees, Oier and Nava, Elvis},
  journal={arXiv preprint arXiv:2512.15692},
  year={2025}
}

@article{an2026feedback,
  title={Feedback World Model Enables Precise Guidance of Diffusion Policy},
  author={An, Tuo and Jia, Jindou and Li, Gen and Li, Jingliang and Zhou, Chuhao and Liu, Pengfei and Lyu, Bofan and Bai, Jiaqi and Guo, Xinying and Li, Geng and others},
  journal={arXiv preprint arXiv:2605.15705},
  year={2026}
}

@article{hou2026world,
  title={World Model for Robot Learning: A Comprehensive Survey},
  author={Hou, Bohan and Li, Gen and Jia, Jindou and An, Tuo and Guo, Xinying and Leng, Sicong and Geng, Haoran and Ze, Yanjie and Harada, Tatsuya and Torr, Philip and others},
  journal={arXiv preprint arXiv:2605.00080},
  year={2026}
}

@inproceedings{bahl2023affordances,
  title={Affordances from human videos as a versatile representation for robotics},
  author={Bahl, Shikhar and Mendonca, Russell and Chen, Lili and Jain, Unnat and Pathak, Deepak},
  booktitle={Proceedings of the IEEE/CVF Conference on Computer Vision and Pattern Recognition},
  pages={13778--13790},
  year={2023}
}

@inproceedings{wu2025afforddp,
  title={Afforddp: Generalizable diffusion policy with transferable affordance},
  author={Wu, Shijie and Zhu, Yihang and Huang, Yunao and Zhu, Kaizhen and Gu, Jiayuan and Yu, Jingyi and Shi, Ye and Wang, Jingya},
  booktitle={Proceedings of the Computer Vision and Pattern Recognition Conference},
  pages={6971--6980},
  year={2025}
}

@article{li2026compassad,
  title={CompassAD: Intent-Driven 3D Affordance Grounding in Functionally Competing Objects},
  author={Li, Jingliang and Jia, Jindou and An, Tuo and Zhou, Chuhao and Chen, Xiangyu and Shan, Shilin and Ma, Boyu and Lyu, Bofan and Li, Gen and Yang, Jianfei},
  journal={arXiv preprint arXiv:2604.02060},
  year={2026}
}

@article{chen2025tool,
  title={Tool-as-interface: Learning robot policies from human tool usage through imitation learning},
  author={Chen, Haonan and Zhu, Cheng and Li, Yunzhu and Driggs-Campbell, Katherine},
  journal={arXiv e-prints},
  pages={arXiv--2504},
  year={2025}
}

@inproceedings{zhulearning,
  title={Learning Generalizable Manipulation Policies with Object-Centric 3D Representations},
  author={Zhu, Yifeng and Jiang, Zhenyu and Stone, Peter and Zhu, Yuke},
  booktitle={Conference on Robot Learning},
  year={2023}
}

@article{chapin2025object,
  title={Object-centric representations improve policy generalization in robot manipulation},
  author={Chapin, Alexandre and Machado, Bruno and Dellandrea, Emmanuel and Chen, Liming},
  journal={arXiv preprint arXiv:2505.11563},
  year={2025}
}

@article{turpin2021gift,
  title={GIFT: Generalizable Interaction-aware Functional Tool Affordances without Labels},
  author={Turpin, Dylan and Wang, Liquan and Tsogkas, Stavros and Dickinson, Sven and Garg, Animesh},
  journal={Robotics: Science and Systems},
  year={2021},
}

@inproceedings{huang2025rekep,
  title={ReKep: Spatio-Temporal Reasoning of Relational Keypoint Constraints for Robotic Manipulation},
  author={Huang, Wenlong and Wang, Chen and Li, Yunzhu and Zhang, Ruohan and Fei-Fei, Li},
  booktitle={Conference on Robot Learning},
  year={2025},
}

@article{wen2023any,
  title={Any-point trajectory modeling for policy learning},
  author={Wen, Chuan and Lin, Xingyu and So, John and Chen, Kai and Dou, Qi and Gao, Yang and Abbeel, Pieter},
  journal={arXiv preprint arXiv:2401.00025},
  year={2023}
}

@inproceedings{he2016deep,
  title={Deep residual learning for image recognition},
  author={He, Kaiming and Zhang, Xiangyu and Ren, Shaoqing and Sun, Jian},
  booktitle={Proceedings of the IEEE conference on computer vision and pattern recognition},
  year={2016}
}

@inproceedings{bertasius2021space,
  title={Is space-time attention all you need for video understanding?},
  author={Bertasius, Gedas and Wang, Heng and Torresani, Lorenzo},
  booktitle={Proceedings of the International Conference on Machine Learning},
  year={2021}
}

@inproceedings{ravi2025sam,
  title={Sam 2: Segment anything in images and videos},
  author={Ravi, Nikhila and Gabeur, Valentin and Hu, Yuan-Ting and Hu, Ronghang and Ryali, Chaitanya and Ma, Tengyu and Khedr, Haitham and R{\"a}dle, Roman and Rolland, Chloe and Gustafson, Laura and others},
  booktitle={International Conference on Learning Representations},
  year={2025}
}

@inproceedings{karaev2024cotracker,
  title={Cotracker: It is better to track together},
  author={Karaev, Nikita and Rocco, Ignacio and Graham, Benjamin and Neverova, Natalia and Vedaldi, Andrea and Rupprecht, Christian},
  booktitle={European conference on computer vision},
  year={2024},
}

@article{chi2025diffusion,
  title={Diffusion policy: Visuomotor policy learning via action diffusion},
  author={Chi, Cheng and Xu, Zhenjia and Feng, Siyuan and Cousineau, Eric and Du, Yilun and Burchfiel, Benjamin and Tedrake, Russ and Song, Shuran},
  journal={The International Journal of Robotics Research},
  year={2025},
}

@inproceedings{lipmanflow,
  title={Flow Matching for Generative Modeling},
  author={Lipman, Yaron and Chen, Ricky TQ and Ben-Hamu, Heli and Nickel, Maximilian and Le, Matthew},
  booktitle={International Conference on Learning Representations},
  year={2023}
}

@article{jia2026physics,
  title={Physics filtering favors the generalization of robot learning},
  author={Jia, Jindou and Han, Shixuan and Wang, Meng and Li, Gen and Yang, Zihan and Zhou, Sicheng and Guo, Kexin and Yang, Jianfei and Yu, Xiang and Wang, Wei and others},
  journal={npj Robotics},
  volume={4},
  number={1},
  pages={48},
  year={2026},
  publisher={Nature Publishing Group UK London}
}

@article{jia2026action,
  title={Action-to-action flow matching},
  author={Jia, Jindou and Li, Gen and Chen, Xiangyu and An, Tuo and Hu, Yuxuan and Li, Jingliang and Guo, Xinying and Yang, Jianfei},
  journal={arXiv preprint arXiv:2602.07322},
  year={2026}
}

@article{bai2026flash,
  title={FLASH: Efficient Visuomotor Policy via Sparse Sampling},
  author={Bai, Jiaqi and Jia, Jindou and Hu, Yuxuan and Li, Gen and Chen, Xiangyu and An, Tuo and Zuo, Kuangji and Yang, Jianfei},
  journal={arXiv preprint arXiv:2605.15492},
  year={2026}
}

\clearpage
\appendix

\phantomsection
\section*{Appendix}

\begin{itemize}
  \item[] \textbf{\ref{app:benchmarks} Benchmark Construction and Implementation Details}
    \begin{itemize}
      \item[] \ref{app:fir_extraction} Extraction of Functional Intermediate Representations
      \item[] \ref{app:mppi_annotation} MPPI-based Demonstration Generation
      \item[] \ref{app:implementation_details} Training and Implementation Details
    \end{itemize}
    
  \item[] \textbf{\ref{app:additional_experimental_results} Additional Experimental Results}
    \begin{itemize}
      \item[] \ref{app:more_qualitative_results_real} More Qualitative Results in the Real World
      \item[] \ref{app:more_qualitative_results_sim} More Qualitative Results in Simulation
    \end{itemize}
\end{itemize}

\section{Benchmark Construction and Implementation Details}
\label{app:benchmarks}

This section provides additional details on benchmark construction and implementation.
We first describe the extraction of different functional intermediate representations in Appx.~\ref{app:fir_extraction}, including those used for representation selection and human-video pretraining.
We then introduce our MPPI-based demonstration generation procedure in Appx.~\ref{app:mppi_annotation}, using the hitting function as a representative example.
Finally, we summarize the training and implementation details of \algoname{} and all baselines in Appx.~\ref{app:implementation_details}.

\begin{table}[ht]
\centering
\caption{\textbf{Statistics of Functional Intermediate Representations for Hitting.}
We report the number of tools, settings, initial configurations, samples per configuration, and total samples used for each representation.}
\label{tab:fir_statistics}

\resizebox{\linewidth}{!}{
\begin{tabular}{lccccc}
\toprule
Representation 
& \# Tools 
& \# Settings 
& \# Initial Configs. 
& \# Samples / Config. 
& \# Total \\
\midrule
Affordance Images                  & 10 & 1 & 1  & 1  & 10  \\
Human Video Prompts                & 7  & 3 & 5  & 5  & 525 \\
Functional Videos                  & 10 & 3 & 30 & 1  & 900 \\
Object Masks                       & 10 & 3 & 30 & 1  & 900 \\
Keypoint Trajectories (Simulation) & 10 & 3 & 30 & 1  & 900 \\
Keypoint Trajectories (Real World) & 7  & 2 & 1  & 20 & 280 \\
\bottomrule
\end{tabular}
}
\end{table}

\subsection{Extraction of Functional Intermediate Representations}
\label{app:fir_extraction}

As discussed in Sec.~3.2, we consider five types of functional intermediate representations: affordance images, human video prompts, functional videos, object masks, and 2D keypoint trajectories. In this section, we use the hitting function as a representative example to illustrate their construction and usage. Specifically, we describe how each representation is extracted and incorporated as an additional condition for the grounded execution policy $\pi_{\mathrm{sys1}}$. The statistics of these representations are summarized in Tab.~\ref{tab:fir_statistics}.

\begin{figure}[ht]
    \centering
    \includegraphics[width=1.0\linewidth]{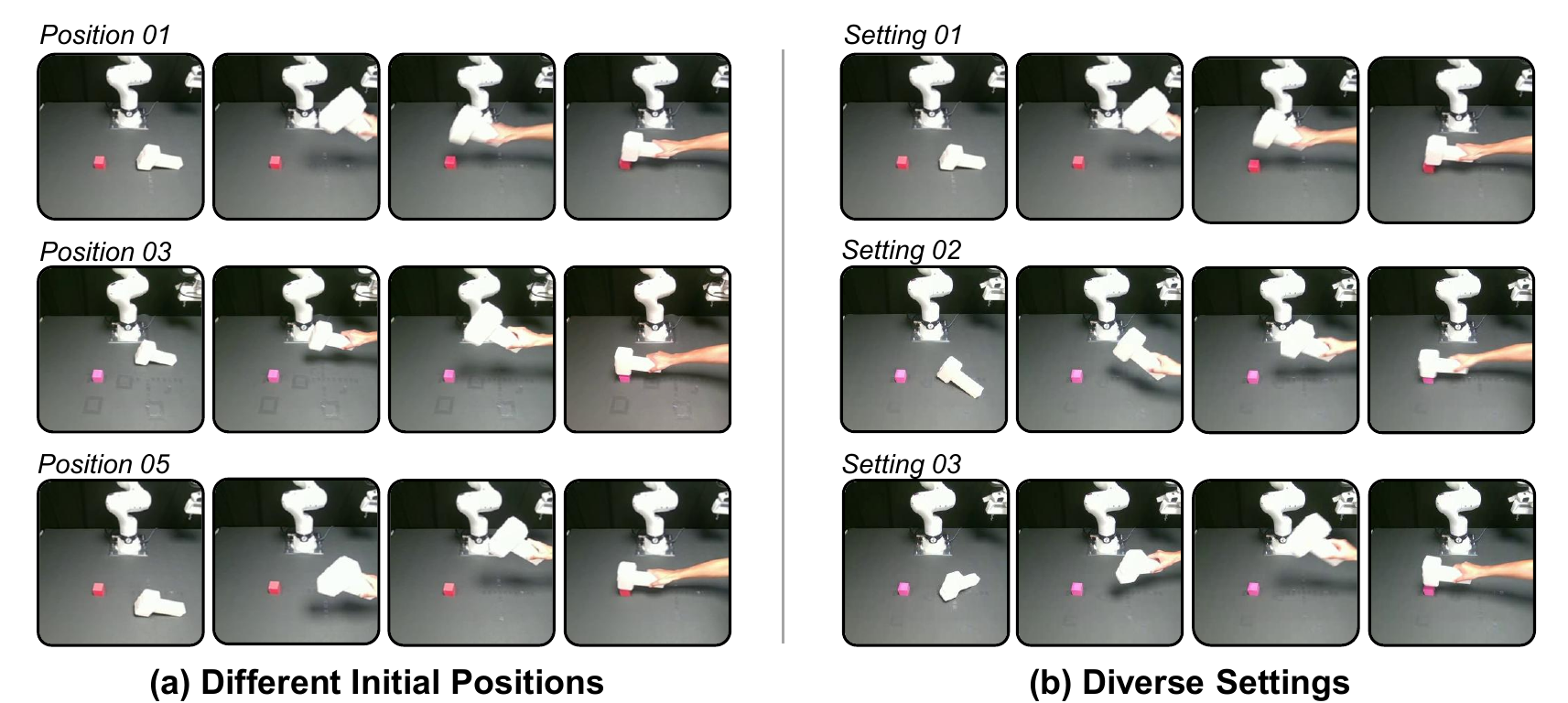}
    \caption{\textbf{Human Video Prompt Samples.}
    (a) Samples from different initial positions. (b) Samples from different settings.}

    \label{fig:human-video-prompts}
\end{figure}

\textbf{Affordance Images.}~~ 
We manually annotate the grasp point and hitting region for each tool, producing 10 affordance images in total. When evaluating functional intermediate representations in Sec.~\ref{sec:Functional Intermediate Representations Selection}, we use the affordance image of the corresponding tool as an additional condition for the grounded execution policy $\pi_{\mathrm{sys1}}$. Since this representation is time-invariant, the same image feature is shared across all time steps of a trajectory.

\textbf{Human Video Prompts.}~~
As shown in Fig.~\ref{fig:human-video-prompts}, we collect human video prompts for all 7 tools. Following the simulation benchmark, each tool has 3 settings. For each setting, we define 5 initial positions and collect 5 videos per position, resulting in 525 human tool-use videos in total. When evaluating functional intermediate representations, we use the video prompt corresponding to the same tool, setting, and sample from Position 01 as an additional condition for the grounded execution policy. These videos provide visual demonstrations of how humans accomplish the same hitting function with different tools. We further validate that human video data can serve as effective action-free pretraining data for functional reasoning in Sec.~\ref{sec:scalability-human-video}.

\textbf{Functional Videos and Object Masks.}~~
For each simulation demonstration, we use the corresponding ground-truth functional execution video as a temporally varying condition.
To construct object masks, we remove the background from each frame while retaining the tool and target object, thereby preserving object-centric geometry and motion while suppressing background appearance.
Both representations are temporally aligned with the robot trajectory and used as sequential conditions for $\pi_{\mathrm{sys1}}$.
With 10 tools, 3 settings, and 30 randomized initial configurations per setting, we obtain 900 functional-video sequences and 900 object-mask sequences in total.

\textbf{Keypoint Trajectories.}~~
We adopt 2D keypoint trajectories as the final functional intermediate representation and extract them for both simulation and real-world data. In simulation, the hitting and target points are directly obtained from the simulator. Given the initial hitting point on the tool, we use SAM2~\citep{ravi2025sam} to segment the corresponding tool mask and sample $N=5$ additional tool keypoints with Farthest Point Sampling (FPS), ensuring spatial coverage of the tool geometry. We then track these sampled tool keypoints with CoTracker~\citep{karaev2024cotracker} to obtain their 2D trajectories. Our simulation benchmark contains 10 tools with 3 settings each. For each setting, we randomly initialize 30 hitting and target positions, resulting in 900 simulation samples.
For real-world data, we manually annotate the target and hitting points for each sample and obtain five additional tool keypoints using the same SAM2- and FPS-based procedure. Unlike simulation, where the hitting and target trajectories are available from the simulator, all 7 real-world keypoints are tracked by CoTracker. The real-world benchmark contains 7 tools, each with 2 hitting patterns
and 1 initial position. We collect 20 samples for each initial position, resulting in 280 real-world samples. 

\subsection{MPPI-based Demonstration Generation}
\label{app:mppi_annotation}

In the simulation benchmark, the action-labeled demonstrations are automatically generated rather than collected through teleoperation.
We use a Model Predictive Path Integral (MPPI) planner with function-specific objectives.
Taking the hitting function as an example, we adopt a four-stage rollout procedure, whose detailed reward terms are summarized in Tab.~\ref{tab:mppi_rewards}.
The stages sequentially encourage the robot to lift the tool, move it toward the target, align the tool-specific hitting point with the target, and execute the final downward strike.

Besides the translational commands optimized by MPPI, we specify a target rotational pose at the end of each stage and generate the corresponding rotational commands through servo control. This helps the tool maintain an appropriate orientation during lifting, alignment, and hitting. Together, the stage-specific MPPI rewards and rotational servo control enable reliable automatic demonstration generation in simulation.

\begin{table}[p]
\centering
\caption{\textbf{Main Reward Terms for MPPI-based Demonstration Generation.}
Here, $d_x$, $d_y$, and $d_{xy}$ denote the distances between the hitting point and the target along the $x$ axis, the $y$ axis, and the $xy$ plane, respectively, while $\Delta d_x$, $\Delta d_y$, and $\Delta d_{xy}$ denote their step-wise reductions.
For the vertical direction, $z_{\mathrm{lift}}$ denotes the tool lifting height from its initial height, $z_1$ denotes the target lifting threshold in Stage~1, $\Delta z_{\mathrm{lift}}$ denotes the step-wise lifting progress, and $e_z$ denotes the error between the current tool height and the target hovering height.
$\Delta z_{\mathrm{head}}$ denotes the vertical displacement of the hitting point between consecutive steps, so $-\Delta z_{\mathrm{head}}$ measures downward motion.
The terms $r_\ast$ reward the current state for satisfying the stage-specific objective, while $r_{\ast\_\mathrm{prog}}$ rewards step-wise progress toward that objective.}

\label{tab:mppi_rewards}
\resizebox{\linewidth}{!}{
\begin{tabular}{lp{0.30\linewidth}p{0.50\linewidth}}
\toprule
Stage & Objective & Main Reward Terms \\
\midrule
Stage 1 
& Lift the tool and align it with the target along the $x$ and $y$ axes.
& 
$
\begin{aligned}
r_{\mathrm{lift}} &= 18.0\min(z_{\mathrm{lift}}, z_1), \\
r_{\mathrm{lift\_prog}} &= 100.0\min(\Delta z_{\mathrm{lift}}, 0.02), \\
r_x &= 22.0\exp(-18.0 d_x), \\
r_{x\_\mathrm{prog}} &= 120.0\min(\Delta d_x, 0.02), \\
r_y &= 8.0\exp(-8.0 d_y), \\
r_{y\_\mathrm{prog}} &= 35.0\min(\Delta d_y, 0.02).
\end{aligned}
$
\\
\midrule
Stage 2 
& Move the tool toward the target while maintaining a suitable hovering height. 
& 
$
\begin{aligned}
r_{xy} &= 14.0\exp(-10.0 d_{xy}), \\
r_x &= 12.0\exp(-18.0 d_x), \\
r_y &= 8.0\exp(-10.0 d_y), \\
r_z &= 7.0\exp(-18.0 e_z), \\
r_{xy\_\mathrm{prog}} &= 95.0\min(\Delta d_{xy}, 0.02), \\
r_{x\_\mathrm{prog}} &= 80.0\min(\Delta d_x, 0.02), \\
r_{y\_\mathrm{prog}} &= 40.0\min(\Delta d_y, 0.02).
\end{aligned}
$
\\
\midrule
Stage 3 
& Finely align the tool-specific hitting point above the target. 
& 
$
\begin{aligned}
r_{xy} &= 14.0\exp(-14.0 d_{xy}), \\
r_x &= 14.0\exp(-24.0 d_x), \\
r_y &= 10.0\exp(-18.0 d_y), \\
r_{xy\_\mathrm{prog}} &= 145.0\min(\Delta d_{xy}, 0.018), \\
r_{x\_\mathrm{prog}} &= 135.0\min(\Delta d_x, 0.018), \\
r_{y\_\mathrm{prog}} &= 110.0\min(\Delta d_y, 0.018), \\
r_z &= 7.0\exp(-18.0 e_z).
\end{aligned}
$
\\
\midrule
Stage 4 
& Keep the hitting point close to the target and execute the downward strike. 
& 
$
\begin{aligned}
r_{xy} &= 12.0\exp(-16.0 d_{xy}), \\
r_x &= 14.0\exp(-28.0 d_x), \\
r_y &= 10.0\exp(-18.0 d_y), \\
r_{xy\_\mathrm{prog}} &= 120.0\min(\Delta d_{xy}, 0.015), \\
r_{x\_\mathrm{prog}} &= 100.0\min(\Delta d_x, 0.015), \\
r_{y\_\mathrm{prog}} &= 80.0\min(\Delta d_y, 0.015), \\
r_{\mathrm{hit}} &= 150.0\min(-\Delta z_{\mathrm{head}}, 0.05),
\end{aligned}
$
\\
\midrule
Contact / Success 
& Encourage valid downward contact with the target.
& 
$
\begin{aligned}
r_{\mathrm{contact}} &= 150.0 \ \text{if valid downward contact}.
\end{aligned}
$
\\
\bottomrule
\end{tabular}
}
\end{table}
\FloatBarrier

\subsection{Training and Implementation Details}
\label{app:implementation_details}

All models are trained on NVIDIA RTX 6000 Pro GPUs. For all methods, we use an action chunk size of 8, a history horizon of 8, and a batch size of 64. We build the simulation benchmark on Robomimic and conduct real-world experiments on a Franka robot platform.

For baseline methods, we train each policy for 120,000 steps using the corresponding action-labeled data. For \algoname{}, we follow the two-stage training procedure described in Sec.~\ref{sec:FORGE}. 
We first pretrain the keypoint motion predictor $\pi_{\mathrm{sys2}}$ on independently collected action-free data for 240,000 steps.
We then freeze the pretrained predictor and fine-tune the grounded execution policy $\pi_{\mathrm{sys1}}$ on the same action-labeled data used by the baselines for 120,000 steps.

\section{Additional Experimental Results}
\label{app:additional_experimental_results}

This section provides additional qualitative comparisons across hitting, sweeping, and hooking in both the real world and simulation. These results complement the quantitative evaluations in the main paper by illustrating representative failure modes of the baselines and how \algoname{} addresses them through function-aware keypoint motion planning.

\begin{figure}[!htbp]
    \centering
    \includegraphics[width=0.82\linewidth]{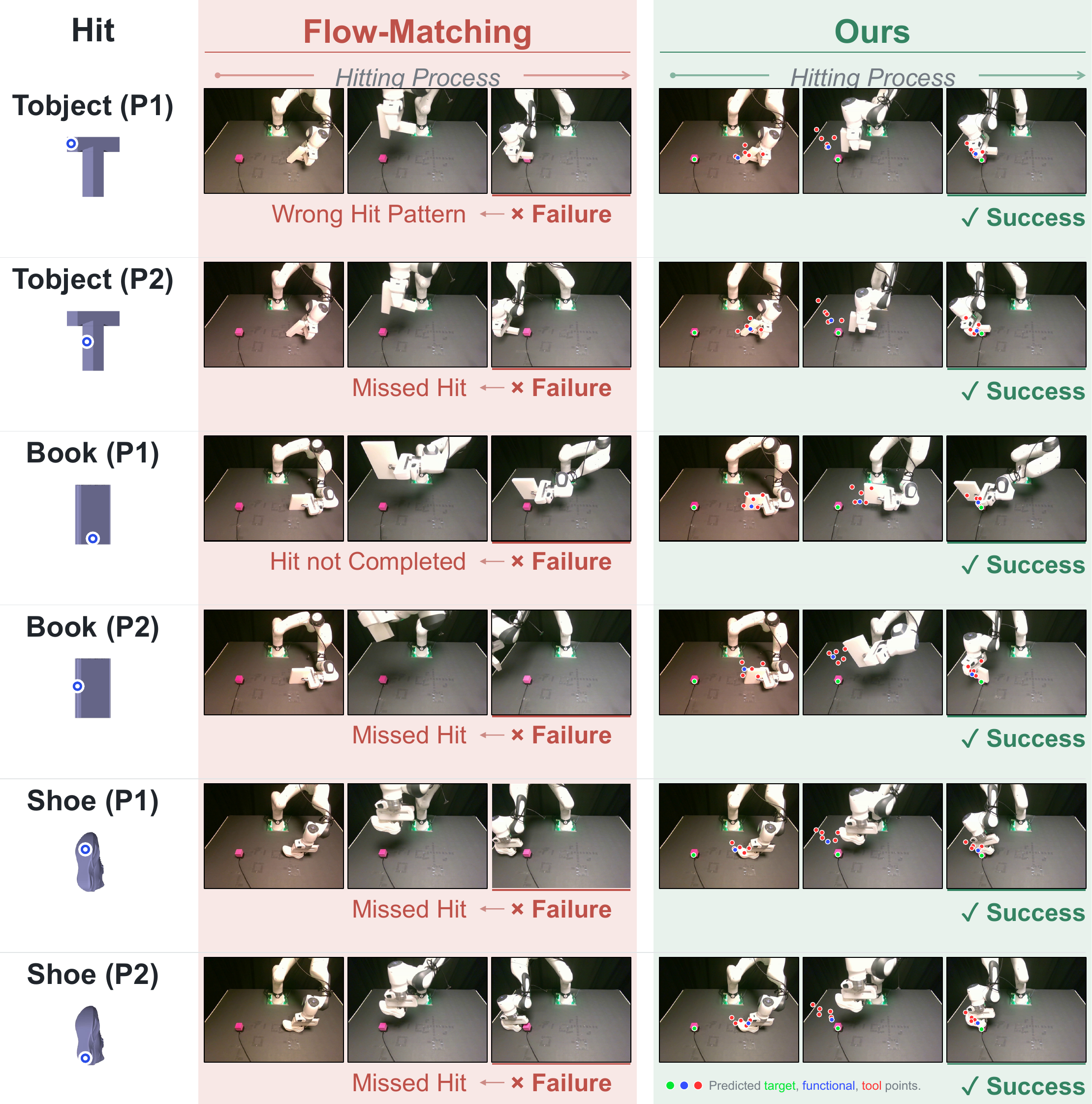}
    \caption{\textbf{Real-World Qualitative Results for Hitting.} Failures of the FM baseline and successful demos of \algoname{} on unseen tools across hitting patterns. The blue point indicates the reference hitting point for each pattern.}
    \label{fig:app-real-hitting}
\end{figure}

\begin{figure}[!htbp]
    \centering
    \includegraphics[width=0.82\linewidth]{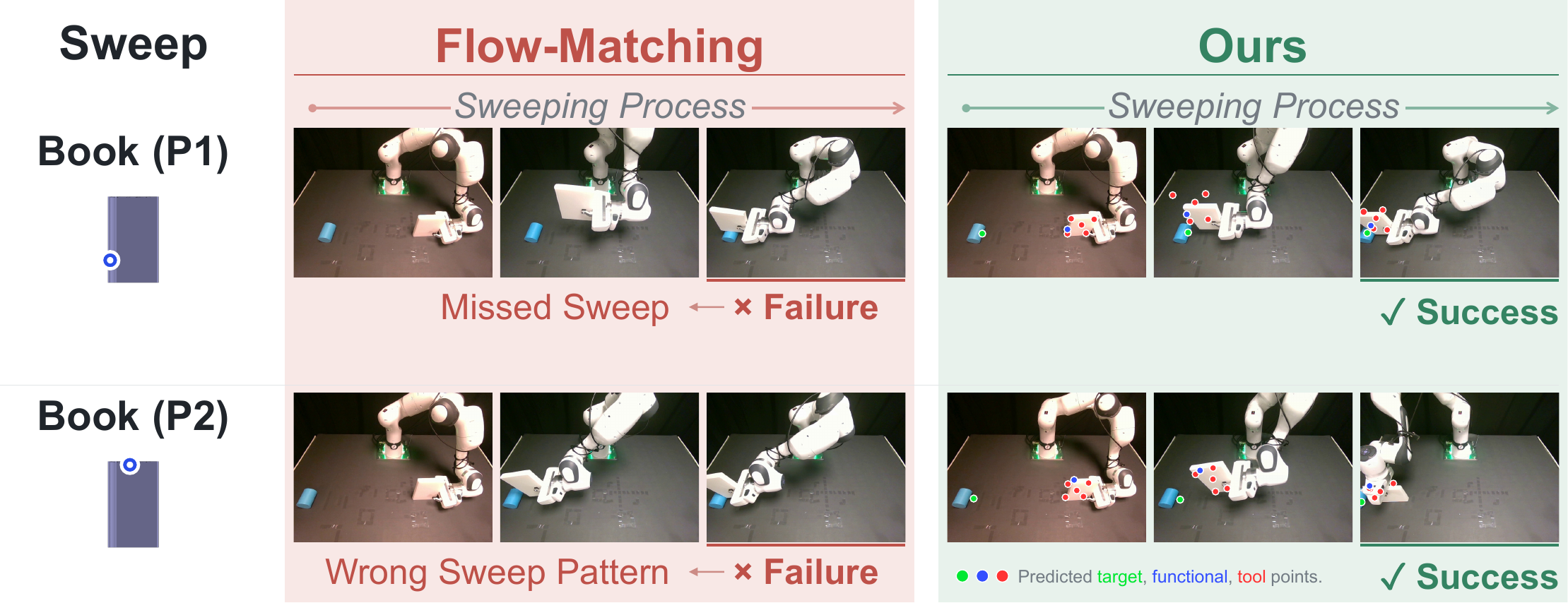}
    \caption{\textbf{Real-World Qualitative Results for Sweeping.} Failures of the FM baseline and successful demos of \algoname{} on unseen tools across sweeping patterns. The blue point indicates the reference sweeping point for each pattern.}
    \label{fig:app-real-sweeping}
\end{figure}

\begin{figure}[!htbp]
    \centering
    \includegraphics[width=0.82\linewidth]{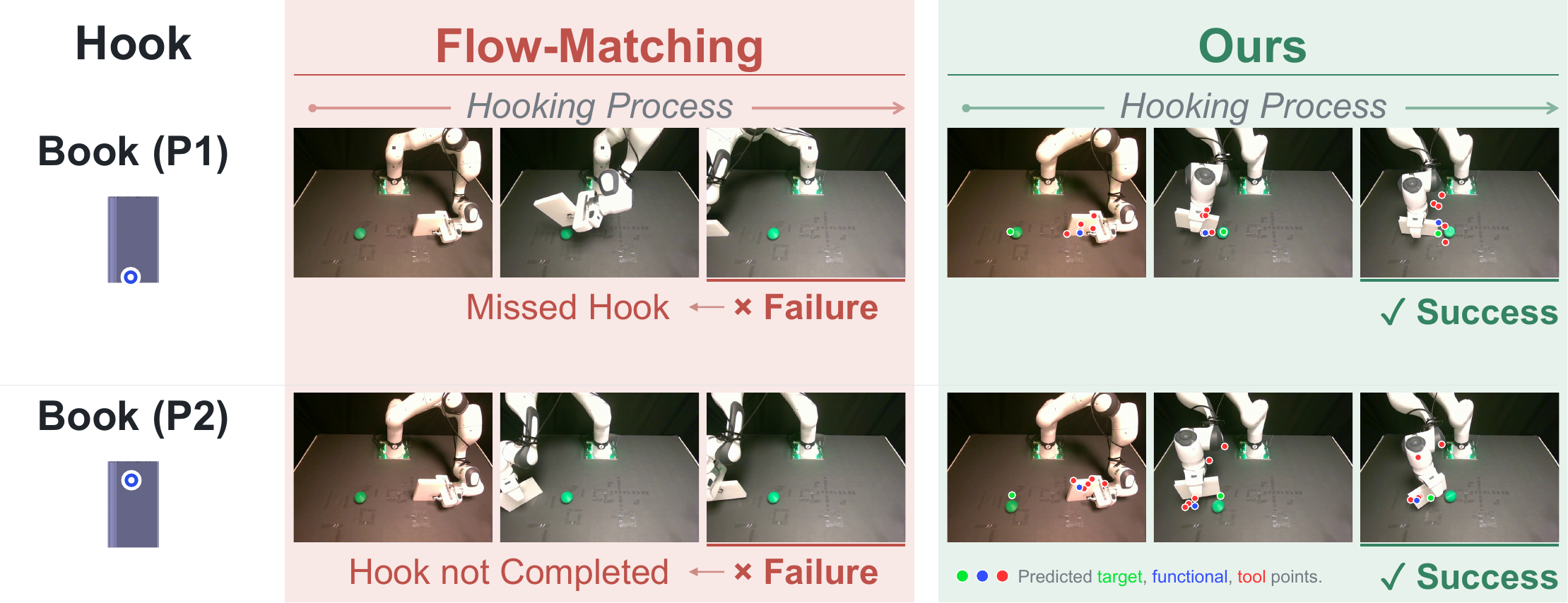}
    \caption{\textbf{Real-World Qualitative Results for Hooking.} Failures of the FM baseline and successful demos of \algoname{} on unseen tools across hooking patterns. The blue point indicates the reference hooking point for each pattern.}
     \label{fig:app-real-hooking}
\end{figure}

\FloatBarrier
\subsection{More Qualitative Results in the Real World}
\label{app:more_qualitative_results_real}

Figs.~\ref{fig:app-real-hitting}--\ref{fig:app-real-hooking} provide additional real-world comparisons between the FM baseline and \algoname{} across hitting, sweeping, and hooking on unseen tools and different functional patterns.
Across the three functions, FM exhibits representative failure modes including missed or incomplete interactions and incorrect functional motion patterns, showing that an end-to-end policy struggles to adapt both the functional region and the required motion to novel tool geometries.
By contrast, \algoname{} uses predicted keypoint-based functional plans to guide functional-region alignment and motion execution, leading to successful interactions across the shown hitting, sweeping, and hooking cases.
These results further demonstrate the benefit of an explicit function-aware intermediate representation for functional generalization.

\begin{figure}[!htbp]
    \centering
    \includegraphics[width=0.82\linewidth]{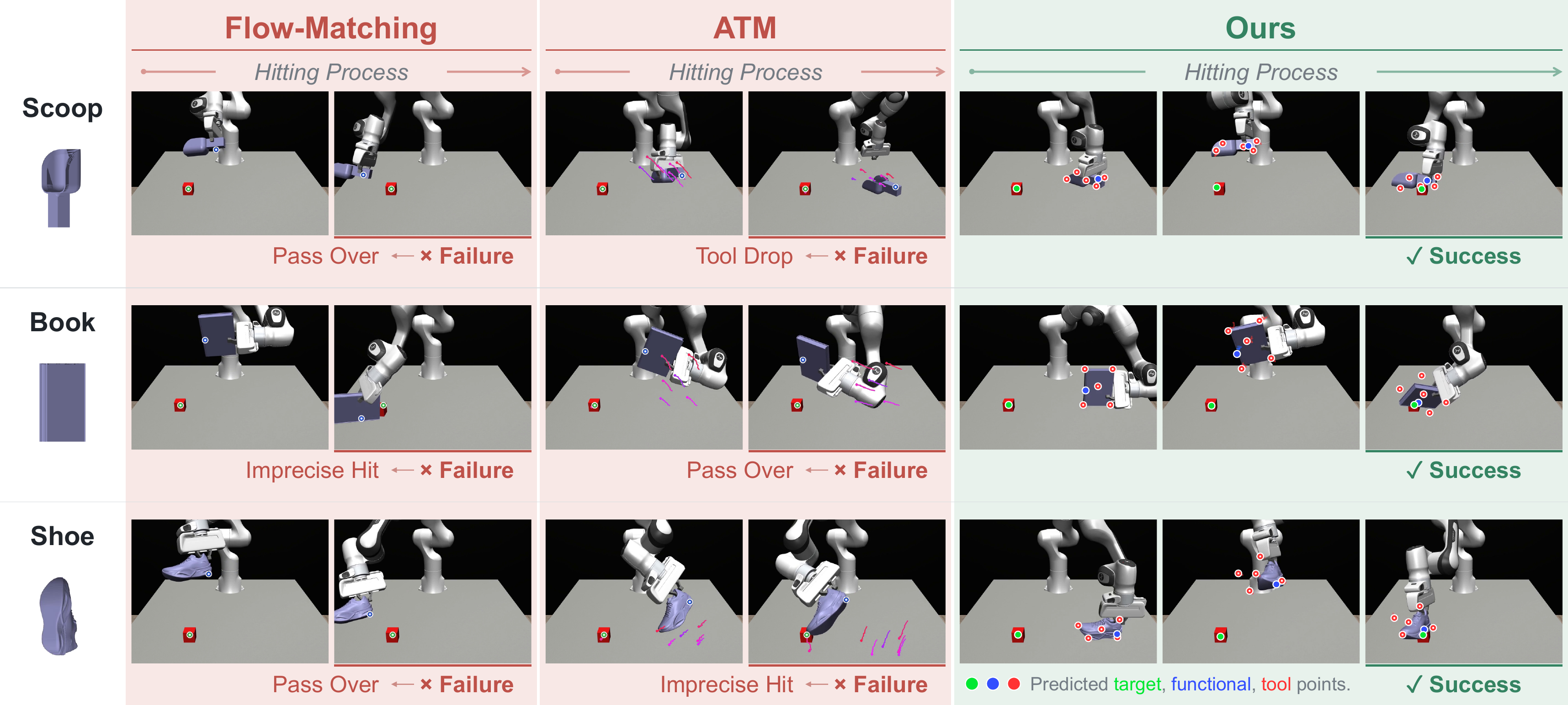}
    \caption{\textbf{Simulation Qualitative Results for Hitting.} Failures of FM and ATM and successful rollouts of \algoname{} on unseen tools.}
    \label{fig:app-hitting-sim}
\end{figure}

\begin{figure}[!htbp]
    \centering
    \includegraphics[width=0.82\linewidth]{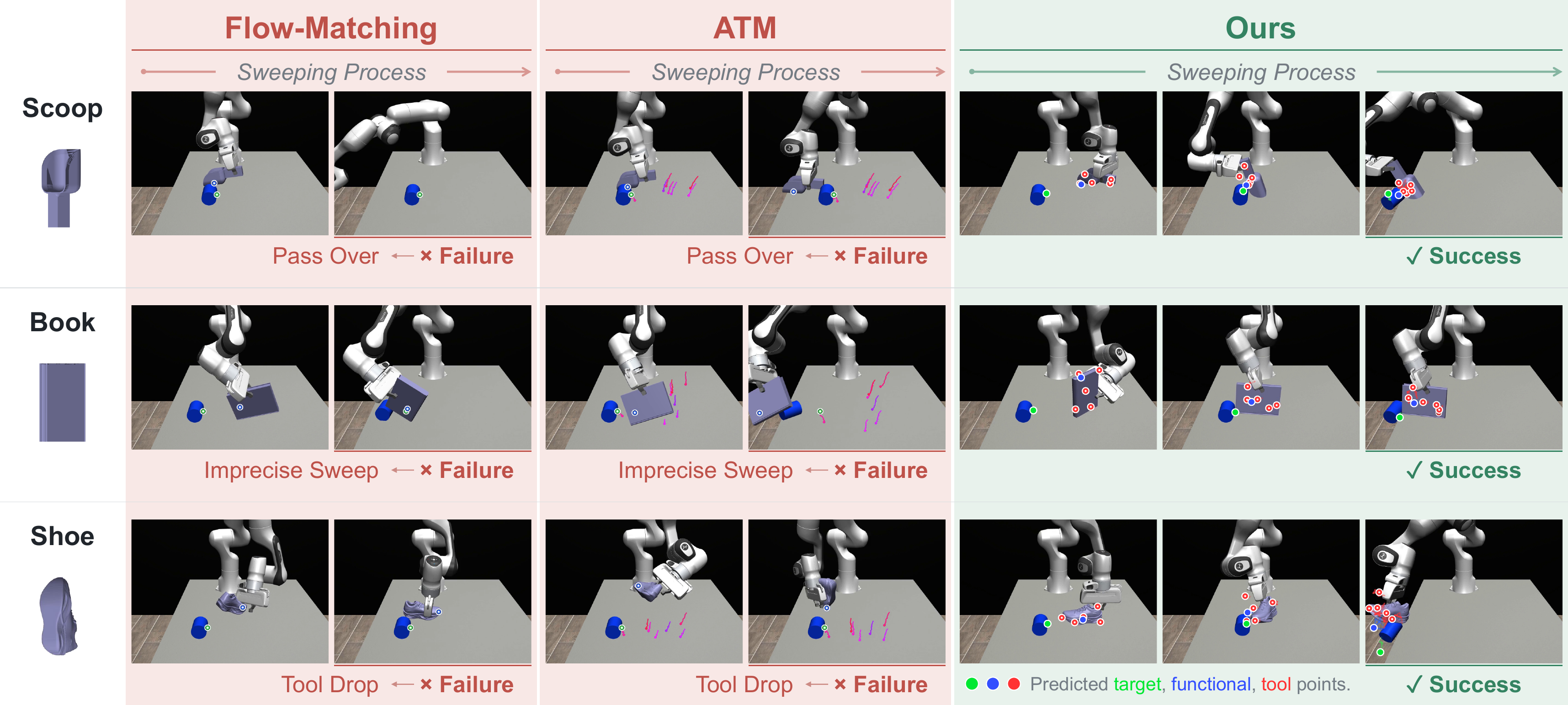}
    \caption{\textbf{Simulation Qualitative Results for Sweeping.} Failures of FM and ATM and successful rollouts of \algoname{} on unseen tools.}
    \label{fig:app-sweeping-sim}
\end{figure}

\begin{figure}[!htbp]
    \centering
    \includegraphics[width=0.82\linewidth]{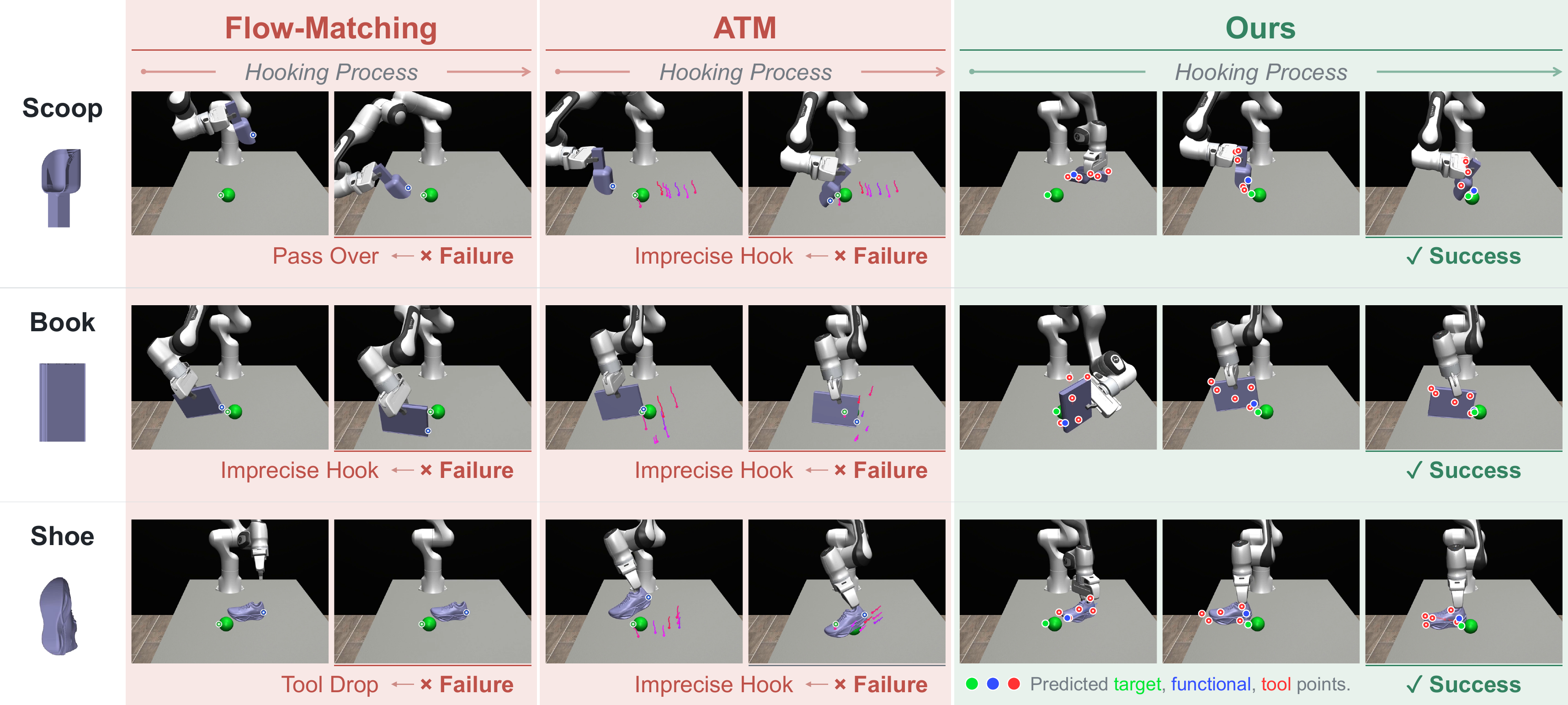}
    \caption{\textbf{Simulation Qualitative Results for Hooking.} Failures of FM and ATM and successful rollouts of \algoname{} on unseen tools.}
    \label{fig:app-hooking-sim}
\end{figure}

\FloatBarrier
\subsection{More Qualitative Results in Simulation}
\label{app:more_qualitative_results_sim}

Figs.~\ref{fig:app-hitting-sim}--\ref{fig:app-hooking-sim} provide additional simulation comparisons among FM, ATM, and \algoname{} across hitting, sweeping, and hooking on unseen tools.
FM exhibits representative failures such as passing over the target, imprecise interactions, and tool dropping.
Although ATM also conditions on keypoint trajectories, its generic trajectories do not explicitly encode function-specific motion intent and can therefore exhibit similar failures.
In contrast, \algoname{} predicts function-aware keypoint trajectories that explicitly guide how the functional region should move toward and interact with the target, enabling more reliable execution across the three functions.

Please refer to our supplementary videos for complete rollouts of both simulation and real-world experiments.

\end{document}